\documentclass{article}

\usepackage[final,nonatbib]{nips_2017}
\usepackage[numbers,compress]{natbib}

\usepackage[utf8]{inputenc} 
\usepackage[T1]{fontenc}    
\usepackage{hyperref}       
\usepackage{url}            
\usepackage{booktabs}       
\usepackage{amsfonts}       
\usepackage{nicefrac}       
\usepackage{microtype}      
\usepackage{placeins}
\usepackage{subfigure}
\usepackage{amsmath}
\usepackage{amssymb}
\usepackage{adjustbox}
\usepackage{multirow}
\usepackage{todonotes}
\usepackage{epstopdf}
\DeclareMathOperator{\tr}{tr}
\newcommand{\calO}{\mathcal{O}}
\newcommand{\bbE}{\mathbb{E}}
\newcommand{\bbR}{\mathbb{R}}

\newcommand\ee[1]{\mathrm{e}{#1}}

\title{Scalable Log Determinants for Gaussian Process Kernel Learning}

\author{
    Kun Dong $^1$,
    David Eriksson $^1$,
    Hannes Nickisch $^2$,
    David Bindel $^1$,
    Andrew Gordon Wilson $^1$ \\
    $^1$ Cornell University, $^2$ Phillips Research Hamburg
}

\graphicspath{{./pics/}}
\begin{document}

\maketitle

\begin{abstract}
For applications as varied as Bayesian neural networks, determinantal
point processes, elliptical graphical models, and
kernel learning for Gaussian processes (GPs),
one must compute a log determinant of an $n \times n$ positive
definite matrix, and its derivatives -- leading to prohibitive
$\mathcal{O}(n^3)$ computations.  We propose novel $\mathcal{O}(n)$
approaches to estimating these quantities from only fast matrix vector
multiplications (MVMs). These stochastic approximations are based on Chebyshev,
Lanczos, and surrogate models, and converge quickly even for kernel matrices
that have challenging spectra.  We leverage these approximations to
develop a scalable Gaussian process approach to kernel learning.
We find that Lanczos is generally superior to Chebyshev for kernel
learning, and that a surrogate approach can be highly efficient and
accurate with popular kernels.

\end{abstract}

\section{Introduction}
\label{sec:introduction}
There is a pressing need for scalable machine learning approaches to extract
rich statistical structure from large datasets.
A common bottleneck ---
arising in determinantal point processes \citep{kulesza2012determinantal},
Bayesian neural networks \citep{mackay1992bayesian},
model comparison \citep{mackay2003information},
graphical models \citep{rue2005gaussian}, and
Gaussian process kernel learning \citep{rasmussen06}
--- is computing a log determinant over a large positive definite matrix.
While we can approximate log determinants by existing stochastic
expansions relying on matrix vector multiplications (MVMs),
these approaches make assumptions, such as near-uniform eigenspectra
\citep{boutsidis2015randomized},
which are unsuitable in machine learning contexts.
For example, the popular RBF kernel gives rise to rapidly decaying
eigenvalues.  Moreover, while standard approaches, such as stochastic
power series, have reasonable asymptotic complexity in the rank of the
matrix, they require too many terms (MVMs) for the precision necessary
in machine learning applications.

Gaussian processes (GPs) provide a principled probabilistic kernel learning
framework, for which a log determinant is of foundational importance.
Specifically, the \emph{marginal likelihood} of a Gaussian process
is the probability of data given only kernel hyper-parameters.
This utility function for kernel learning compartmentalizes into
automatically calibrated model fit and complexity terms --- called
\emph{automatic Occam's razor} --- such that the simplest models
which explain the data are automatically favoured
\citep{rasmussen01, rasmussen06}, without the need for
approaches such as cross-validation, or regularization, which can be
costly, heuristic, and involve substantial hand-tuning and human
intervention.
The automatic complexity penalty, called the \emph{Occam's factor}
\citep{mackay2003information},
is a log determinant of a kernel (covariance) matrix, related
to the volume of solutions that can be expressed by the Gaussian
process.

Many current approaches to scalable Gaussian processes
\cite[e.g.,][]{quinonero2005unifying, le2013fastfood, hensman2013uai}
focus on inference assuming a fixed kernel, or use approximations that
do not allow for very flexible kernel learning \citep{wilson2014thesis},
due to poor scaling with number of basis functions or inducing points.
Alternatively, approaches which exploit algebraic structure in kernel
matrices can provide highly expressive kernel learning \citep{wilson2014fast},
but are essentially limited to grid structured data.

Recently, \citet{wilson2015kernel} proposed the {\em structured kernel interpolation}
(SKI) framework, which generalizes structuring exploiting methods to arbitrarily
located data.  SKI works by providing accurate and fast matrix vector multiplies
(MVMs) with kernel matrices, which can then be used in iterative solvers such
as linear conjugate gradients for scalable GP inference.  However, evaluating
the marginal likelihood and its derivatives, for kernel learning, has followed a
scaled eigenvalue approach \citep{wilson2014fast, wilson2015kernel} instead
of iterative MVM approaches.  This approach can be inaccurate, and relies on
a fast eigendecomposition of a structured matrix, which is not available in many
consequential situations where fast MVMs are available, including: (i) additive covariance
functions, (ii) multi-task learning, (iii) change-points \citep{herlands2016scalable}, and (iv) diagonal
corrections to kernel approximations \citep{snelson2006sparse}.  Fiedler
\citep{fiedler84hankelLoewner} and Weyl \citep{weyl1912} bounds have
been used to extend the scaled eigenvalue approach \citep{flaxman2015fast, herlands2016scalable},
but are similarly limited.  These extensions are often very approximate, and do not apply beyond
sums of two and three matrices,
where each matrix in the sum must have a fast eigendecomposition.

In machine learning there has recently been renewed interest in MVM based approaches to
approximating log determinants, such as the Chebyshev \citep{han2015large}
and Lanczos \citep{ubarufast} based methods, although these approaches go
back at least two decades in quantum chemistry computations
\citep{bai1998computing}.
Independently, several authors have proposed various methods to compute
derivatives of log determinants~\citep{mackay1997efficient,stein2013stochastic}.
But {\em both} the log determinant {\em and} the derivatives
are needed for efficient GP marginal likelihood learning:
the derivatives are required for gradient-based optimization,
while the log determinant itself is needed for model comparison,
comparisons between the likelihoods at local maximizers,
and fast and effective choices of starting points and step sizes
in a gradient-based optimization algorithm.

In this paper, we develop novel scalable and general purpose
Chebyshev, Lanczos, and surrogate approaches for efficiently
and accurately computing
both the log determinant and its derivatives
simultaneously.
Our methods use only fast MVMs, and re-use the
same MVMs for both computations.
In particular:

\vspace{-1mm}
\begin{itemize}
    \item We derive fast methods for simultaneously computing the log
          determinant and its derivatives by stochastic Chebyshev,
          stochastic Lanczos, and surrogate models, from MVMs alone.
          We also perform an error analysis and extend these approaches to
          higher order derivatives.

    \item These methods enable fast
    GP kernel learning whenever fast MVMs are possible, including
    applications where alternatives such as scaled eigenvalue
    methods (which rely on fast eigendecompositions) are not,
    such as for (i) diagonal
    corrections for better kernel approximations, (ii) additive covariances, (iii) multi-task approaches,
    and (iv) non-Gaussian likelihoods.

    \item We illustrate the performance of our approach on several large,
          multi-dimensional datasets, including a consequential crime prediction
          problem, and a precipitation problem with $n = 528,474$ training points.
          We consider a variety of kernels, including deep kernels \citep{wilson2016deep},
          diagonal corrections,
          and both Gaussian and non-Gaussian
          likelihoods.
    \item We have released code and tutorials as an
    extension to the GPML
    library \citep{rasmussen10gpml} at
    \url{https://github.com/kd383/GPML_SLD}.
    A Python implementation of our approach
    is also available through the GPyTorch library:
    \url{https://github.com/jrg365/gpytorch}.
\end{itemize}
\vspace{-1mm}
When using our approach in
conjunction with SKI \citep{wilson2015kernel}
for fast MVMs, GP kernel learning is $\mathcal{O}(n + g(m))$, for $m$ inducing
points and $n$ training points, where
$g(m) \leq m \log m$. With algebraic approaches such as SKI we also do not need
to worry about quadratic storage in inducing points, since symmetric Toeplitz
and Kronecker matrices can be stored with at most linear cost, without needing
to explicitly construct a matrix.

Although we here use SKI for fast
MVMs, we emphasize that the proposed iterative
approaches are generally applicable, and can easily be used in conjunction with
\emph{any} method that admits fast MVMs, including classical
inducing point methods \citep{quinonero2005unifying}, finite basis
expansions \citep{le2013fastfood}, and the popular stochastic variational
approaches \citep{hensman2013uai}.  Moreover, stochastic variational
approaches can naturally be combined with SKI to further accelerate
MVMs \citep{wilson2016stochastic}.

We start in \S \ref{sec:background} with an introduction to GPs
and kernel approximations.  In \S \ref{sec:methods} we introduce
stochastic trace
estimation and Chebyshev (\S \ref{sec:cheb}) and
Lanczos (\S \ref{sec:lanc}) approximations.
In \S \ref{sec:properties}, we describe the different sources of error
in our approximations.
In \S \ref{sec:experiments} we consider experiments on
several large real-world data sets.
We conclude in \S \ref{sec:discussion}.  The supplementary
materials also contain several additional experiments and
details.

\section{Background}
\label{sec:background}
A Gaussian process (GP) is a collection of random variables,
any finite number of which
have a joint Gaussian distribution \citep[e.g.,][]{rasmussen06}.
A GP can be used to define a
distribution over functions $f(x) \sim \mathcal{GP}(\mu(x),k(x,x'))$,
where each function value is a random variable indexed by $x \in \bbR^d$,
and $\mu : \bbR^d \rightarrow \bbR$ and $k : \bbR^d \times \bbR^d \rightarrow \bbR$
are the mean and covariance functions of the process.

The covariance function is often chosen to be an RBF or Mat\'ern kernel (see the supplementary
material for more details).
We denote any kernel hyperparameters by the vector $\theta$. To be
concise we will generally not explicitly denote the dependence of $k$
and associated matrices on $\theta$.

For any locations $X = \{x_1,\ldots,x_n\} \subset \bbR^d$,
$f_X \sim \mathcal{N}(\mu_X, K_{XX})$
where $f_X$ and $\mu_X$ represent the vectors of function values for $f$ and $\mu$
evaluated at each of the $x_i \in X$, and $K_{XX}$ is the matrix whose $(i,j)$
entry is $k(x_i, x_j)$.  Suppose we have a vector of corresponding
function values $y \in \bbR^n$, where each entry is contaminated
by independent Gaussian noise with variance $\sigma^2$.
Under a Gaussian process prior depending on the covariance hyperparameters
$\theta$, the log marginal likelihood is given by
\begin{equation}
    \label{eq:mloglik}
    \mathcal{L}(\theta|y) =
    -\frac{1}{2}\left[(y-\mu_X)^T\alpha + \log |\tilde{K}_{XX}| + n\log 2\pi\right]
\end{equation}
where $\alpha = \tilde{K}_{XX}^{-1}(y-\mu_X)$ and
$\tilde{K}_{XX} = K_{XX} + \sigma^2 I$.
Optimization of (\ref{eq:mloglik}) is expensive, since the cheapest way of
evaluating $\log |\tilde{K}_{XX}|$ and its derivatives without taking advantage of the
structure of $\tilde{K}_{XX}$ involves computing the $\mathcal{O}(n^3)$ Cholesky
factorization of $\tilde{K}_{XX}$.
$\calO(n^3)$ computations is too expensive for inference and learning beyond
even just a few thousand points.

A popular approach to GP scalability is to replace the exact kernel
$k(x,z)$ by an approximate kernel that admits fast computations \cite{quinonero2005unifying}.
Several methods approximate $k(x,z)$ via {\em inducing points}
$U = \{u_j\}_{j=1}^m \subset \bbR^d$.
An example is the subset of regressor (SoR) kernel:
\[
    k^{SoR}(x,z) = K_{xU} K_{UU}^{-1} K_{Uz}
\]
which is a low-rank approximation \cite{silverman1985some}.
The SoR matrix $K^{\mathrm{SoR}}_{XX} \in \bbR^{n \times n}$
has rank at most $m$,
allowing us to solve linear systems involving
$\tilde{K}^{\mathrm{SoR}}_{XX} = K^{\mathrm{SoR}}_{XX} + \sigma^2 I$ and
to compute $\log |\tilde{K}^{\mathrm{SoR}}_{XX}|$
in $\calO(m^2 n + m^3)$ time.
Another popular
kernel approximation is the fully independent training conditional
(FITC), which is a diagonal correction of SoR so that the diagonal is the
same as for the original kernel \cite{snelson2006sparse}.  Thus kernel
matrices from FITC have low-rank plus diagonal structure.
This modification has had exceptional practical significance, leading
to improved point predictions and much more realistic predictive
uncertainty \cite{quinonero2005unifying, quinonero2007}, making FITC
arguably the most popular approach for scalable Gaussian processes.

\citet{wilson2015kernel} provides a mechanism for fast MVMs through
proposing the structured kernel interpolation (SKI) approximation,
\begin{align}
    K_{XX} \approx W K_{UU} W^{T}
    \label{eqn: ski}
\end{align}
where $W$ is an $n$-by-$m$ matrix of interpolation weights;
the authors of~\cite{wilson2015kernel} use local cubic interpolation
so that $W$ is sparse.
The sparsity in $W$ makes it possible to naturally exploit algebraic
structure (such as Kronecker or Toeplitz structure) in $K_{UU}$ when
the inducing points $U$
are on a grid, for extremely fast matrix vector multiplications with
the approximate $K_{XX}$
even if the data inputs $X$ are arbitrarily located.
For instance, if $K_{UU}$ is Toeplitz, then each MVM with the
approximate $K_{XX}$ costs only $\calO(n + m \log m)$.
By contrast, placing the inducing points $U$ on a grid for classical inducing
point methods, such as SoR or FITC, does
not result in substantial performance gains, due to the costly
cross-covariance matrices $K_{xU}$ and $K_{Uz}$.

\section{Methods}
\label{sec:methods}
Our goal is to estimate, for a symmetric positive definite matrix $\tilde K$,
\[
  \log |\tilde K| = \tr(\log(\tilde K)) \quad \mbox{and} \quad
  \frac{\partial}{\partial \theta_i} \left[\log |\tilde K|\right] =
   \tr\left( \tilde K^{-1} \left(\frac{\partial \tilde
  K}{\partial \theta_i} \right) \right),
\]
where $\log$ is the matrix logarithm \cite{higham2008functions}.
We compute the traces involved in both the log determinant and its
derivative via
{\em stochastic trace estimators}~\cite{hutchinson1990stochastic},
which approximate the trace of a matrix using only matrix vector products.

The key idea is that for a given matrix $A$ and
a random probe vector $z$ with independent entries with mean zero
and variance one, then $\tr(A) = \bbE[z^T A z]$;
a common choice is to let the entries of the probe vectors be Rademacher
random variables.
In practice, we estimate the trace by the sample mean over $n_z$
independent probe vectors.  Often
surprisingly few probe vectors suffice.

To estimate $\tr(\log(\tilde K))$, we need to multiply
$\log(\tilde K)$ by probe vectors.
We consider two ways to estimate $\log(\tilde K) z$: by a polynomial
approximation of $\log$ or by using the connection between the
Gaussian quadrature rule and the Lanczos
method \cite{han2015large,ubarufast}.
In both cases, we show how to re-use the same probe vectors for an
inexpensive coupled estimator of the derivatives.
In addition, we may use standard radial basis function interpolation
of the log determinant evaluated at a few systematically
chosen points in the hyperparameter space as an inexpensive
surrogate for the log determinant.

\subsection{Chebyshev}
\label{sec:cheb}

Chebyshev polynomials are defined by the recursion
\[
  T_0(x) = 1, \quad T_1(x) = x, \quad
  T_{j+1}(x) = 2xT_j(x)-T_{j-1}(x)\; \text{ for }\; j\geq 1.
\]
For $f : [-1,1] \to \bbR$ the Chebyshev interpolant of degree $m$ is
\[
  f(x) \approx p_m(x) := \sum_{j=0}^m c_j T_j(x), \quad
  \mbox{where }
  c_j =
    \frac{2 - \delta_{j0}}{m+1} \displaystyle\sum_{k=0}^m f(x_k)T_j(x_k)
\]
where $\delta_{j0}$ is the Kronecker delta and
$x_k = \cos(\pi(k+1/2)/(m+1))$
for $k=0,1,2,\ldots,m$; see~\cite{gil2007numerical}.
Using the Chebyshev interpolant of $\log(1+\alpha x)$,
we approximate $\log |\tilde K|$ by
\begin{align*}
  \log|\tilde K| -n\log\beta  &= \log|I+\alpha B|
  \approx \sum_{j=0}^m c_j \tr(T_j(B))
\end{align*}
when $B=(\tilde K/\beta-1)/\alpha$ has eigenvalues
$\lambda_i \in (-1,1)$.

For stochastic estimation of $\tr(T_j(B))$, we only need to compute
$z^T T_j(B)z$ for each given probe vector $z$. We compute
vectors $w_j=T_j(B)z$ and $\partial w_j/\partial \theta_i$
via the coupled recurrences
\[
\begin{aligned}
  w_0 &= z, &
  w_1 &= Bz, &
  w_{j+1} &= 2Bw_j - w_{j-1} \mbox{ for } j \geq 1, \\
  \frac{\partial w_0}{\partial \theta_i} &= 0, &
  \frac{\partial w_1}{\partial \theta_i} &= \frac{\partial B}{\partial \theta_i} z, &
  \frac{\partial w_{j+1}}{\partial \theta_i} &=
  2\left(
    \frac{\partial B}{\partial \theta_i} w_j +
    B \frac{\partial w_j}{\partial \theta_i}
  \right) -
  \frac{\partial w_{j-1}}{\partial \theta_i} \mbox{ for } j \geq 1.
\end{aligned}
\]
This gives the estimators
\[
  \log |\tilde K| \approx \mathbb{E}\left[\sum_{j=0}^m c_j
  z^Tw_j\right]
  \quad \mbox{ and } \quad
  \frac{\partial}{\partial \theta_i} \log|\tilde K| \approx
  \mathbb{E}\left[\sum_{j=0}^m c_j z^T\frac{\partial w_j}{\partial \theta_i}\right].
\]
Thus, each derivative of the approximation costs two extra MVMs per term.

\subsection{Lanczos}
\label{sec:lanc}

We can also approximate $z^T \log(\tilde K) z$ via a Lanczos
decomposition; see~\cite{golub2010matrices} for discussion
of a Lanczos-based computation of $z^T f(\tilde K) z$
and~\cite{ubarufast,bai1998computing} for stochastic Lanczos estimation of
log determinants.  We run $m$
steps of the Lanczos algorithm, which computes the decomposition
\[
  \tilde K Q_m = Q_m T + \beta_m q_{m+1} e_m^T
\]
where
$Q_m = \begin{bmatrix}
q_1 & q_2 & \ldots q_m
\end{bmatrix} \in \bbR^{n \times m}$
is a matrix with orthonormal columns such that $q_1 = z/\|z\|$,
$T \in \bbR^{m \times m}$ is tridiagonal, $\beta_m$ is the residual, and
$e_m$ is the $m$th Cartesian unit vector.  We estimate
\begin{equation} \label{eq:gaussq}
  z^T f(\tilde K) z \approx
  e_1^T f(\|z\|^2 T) e_1
\end{equation}
where $e_1$ is the first column of the identity. The Lanczos algorithm
is numerically unstable. Several practical implementations resolve this issue
\cite{cullum2002lanczos,saad1992numerical}.
The approximation~\eqref{eq:gaussq}
corresponds to a Gauss quadrature rule for the Riemann-Stieltjes
integral of the measure associated with the eigenvalue distribution
of $\tilde K$.  It is exact when $f$ is a polynomial of degree up
to $2m-1$.  This approximation is also exact when $\tilde K$ has at most $m$ distinct
eigenvalues, which is particularly relevant to Gaussian process
regression, since frequently the kernel matrices only have a small
number of eigenvalues that are not close to zero.

The Lanczos decomposition also allows us to estimate derivatives
of the log determinant at minimal cost.
Via the Lanczos decomposition, we have
\[
  \hat{g}
  = Q_m (T^{-1} e_1 \|z\|)
  \approx \tilde K^{-1}z.
\]
This approximation requires no additional matrix vector multiplications
beyond those used to compute the Lanczos decomposition, which we already
used to estimate $\log(\tilde K) z$; in exact arithmetic, this is
equivalent to $m$ steps of CG.
Computing $\hat{g}$
in this way takes $\calO(mn)$ additional time; subsequently, we only need
one matrix-vector multiply by $\partial \tilde K / \partial \theta_i$
for each probe vector to estimate
$\tr(\tilde K^{-1} (\partial \tilde K / \partial \theta_i)) =
 \mathbb{E}[(\tilde K^{-1} z)^T (\partial \tilde K/\partial \theta_i) z]$.

\subsection{Diagonal correction to SKI}
\label{sup:diagcorrection}

The SKI approximation may provide a poor estimate of the diagonal entries
of the original kernel matrix for kernels with limited smoothness, such as the
Mat\'ern kernel.  In general, diagonal corrections to scalable kernel approximations
can lead to great performance gains.  Indeed, the popular FITC method
\citep{snelson2006sparse} is exactly a diagonal correction of subset of
regressors (SoR).

We thus modify the SKI approximation to add a diagonal matrix $D$,
\begin{align}
    K_{XX} \approx W K_{UU} W^T + D \,,
    \label{eqn: diagcorr}
\end{align}
such that the diagonal of the approximated $K_{XX}$ is exact.  In other words,
$D$ substracts the diagonal of $W K_{UU} W^T$ and adds the true diagonal of $K_{XX}$.  This
modification is not possible for the scaled eigenvalue method for approximating log determinants in
\citep{wilson2015kernel}, since adding a diagonal matrix
makes it impossible to approximate the eigenvalues of $K_{XX}$ from the eigenvalues of $K_{UU}$.

However, Eq.~\eqref{eqn: diagcorr} still admits fast MVMs and thus works with our approach
for estimating the log determinant and its derivatives.  Computing $D$ with SKI costs only $\calO(n)$ flops since $W$ is sparse for local cubic interpolation. We can therefore compute
$(W^Te_i)^TK_{UU}(W^Te_i)$ in $\calO(1)$ flops.

\subsection{Estimating higher derivatives}
\label{sec:higherderivatives}

We have already described how to use stochastic estimators to compute
the log marginal likelihood and its first derivatives. The same approach applies
to computing higher-order derivatives for a Newton-like iteration,
to understand the sensitivity of the maximum likelihood parameters,
or for similar tasks.
The first derivatives of the full log marginal likelihood are
\[
  \frac{\partial \mathcal{L}}{\partial \theta_i} =
  -\frac{1}{2} \left[
  \tr\left(
    \tilde K^{-1} \frac{\partial \tilde K}{\partial \theta_i}
  \right) - \alpha^T \frac{\partial \tilde K}{\partial \theta_i} \alpha
  \right]
\]
and the second derivatives of the two terms are
\begin{align*}
  \frac{\partial^2}{\partial \theta_i \partial \theta_j}
  \left[ \log |\tilde K| \right] &=
  \tr\left(
    \tilde K^{-1} \frac{\partial^2 \tilde K}
                      {\partial \theta_i \partial \theta_j} -
    \tilde K^{-1} \frac{\partial \tilde K}{\partial \theta_i}
    \tilde K^{-1} \frac{\partial \tilde K}{\partial \theta_j}
  \right), \\
  \frac{\partial^2}{\partial \theta_i \partial \theta_j}
  \left[ (y-\mu_X)^T \alpha \right] &=
    2
    \alpha^T \frac{\partial \tilde K}{\partial \theta_i}
    \tilde K^{-1}
    \frac{\partial \tilde K}{\partial \theta_j} \alpha -
    \alpha^T
    \frac{\partial^2 \tilde K}
         {\partial \theta_i \partial \theta_j}
    \alpha.
\end{align*}
Superficially, evaluating the second derivatives would appear to
require several additional solves above and beyond those used to
estimate the first derivatives of the log determinant.  In fact,
we can get an unbiased estimator for the second derivatives with no
additional solves, but only fast products with the derivatives of
the kernel matrices.  Let $z$ and $w$ be independent probe vectors,
and define $g = \tilde K^{-1} z$ and $h = \tilde K^{-1} w$.  Then
\begin{align*}
  \frac{\partial^2}{\partial \theta_i \partial \theta_j}
  \left[ \log |\tilde K| \right] &=
  \mathbb{E}
  \left[
    g^T \frac{\partial^2 \tilde K}
             {\partial \theta_i \partial \theta_j} z -
    \left( g^T \frac{\partial \tilde K}{\partial \theta_i} w \right)
    \left( h^T \frac{\partial \tilde K}{\partial \theta_j} z \right)
  \right], \\
  \frac{\partial^2}{\partial \theta_i \partial \theta_j}
  \left[ (y-\mu_X)^T \alpha \right] &=
    2 \mathbb{E} \left[
    \left( z^T \frac{\partial \tilde K}{\partial \theta_i} \alpha \right)
    \left( g^T \frac{\partial \tilde K}{\partial \theta_j} \alpha \right)
    \right] -
    \alpha^T
    \frac{\partial^2 \tilde K}
         {\partial \theta_i \partial \theta_j}
    \alpha.
\end{align*}
Hence, if we use the stochastic Lanczos method to compute the log
determinant and its derivatives, the additional work required to
obtain a second derivative estimate is one MVM by each second
partial of the kernel for each probe vector and for $\alpha$,
one MVM of each first partial of the kernel with $\alpha$,
and a few dot products.

\subsection{Radial basis functions}
\label{sec:rbf_interp}
Another way to deal with the log determinant and its derivatives
is to evaluate the log determinant term at a few
systematically chosen points in the space of hyperparameters
and fit an interpolation approximation to these values.  This
is particularly useful
when the kernel depends on a modest number of hyperparameters
(e.g., half a dozen), and thus the number of points we need to
precompute is relatively small.
We refer to this method as a surrogate, since
it provides an inexpensive substitute for the log determinant and its derivatives.
For our surrogate approach, we use radial basis function (RBF)
interpolation with a cubic kernel and a linear tail. See
e.g.~\cite{buhmann2000radial,fasshauer2007meshfree,schaback2006kernel,wendland2004scattered}
and the supplementary material for more details on RBF interpolation.

\section{Error properties}
\label{sec:properties}
In addition to the usual errors from sources such as solver
termination criteria and floating point arithmetic, our approach to
kernel learning involves several additional sources of error:
we approximate the true kernel with one that enables fast MVMs,
we approximate traces using stochastic estimation,
and we approximate the actions of $\log(\tilde K)$ and
$\tilde K^{-1}$ on probe vectors.

We can compute first-order estimates of the sensitivity of the
log likelihood to perturbations in the kernel using the same
stochastic estimators we use for the derivatives
with respect to hyperparameters.  For example, if $\mathcal{L}^{\mathrm{ref}}$
is the likelihood for a reference kernel
$\tilde K^{\mathrm{ref}} = \tilde K + E$, then
\[
    \mathcal{L}^{\mathrm{ref}}(\theta|y) =
    \mathcal{L}(\theta|y)
    -\frac{1}{2} \left( \mathbb{E}\left[g^T Ez\right] - \alpha^T E \alpha \right)
    + O(\|E\|^2),
\]
and we can bound the change in likelihood at first order by
$\|E\|\left( \|g\| \|z\| + \|\alpha\|^2 \right)$.
Given bounds on the norms of $\partial E/\partial \theta_i$, we can
similarly estimate changes in the gradient of the likelihood,
allowing us to bound how the marginal likelihood hyperparameter
estimates depend on kernel approximations.

If $\tilde{K} = U \Lambda U^T + \sigma^2 I$,
the Hutchinson trace estimator has known variance~\cite{Avron:2011:Randomized}
\[
    \operatorname{Var}[z^T \log(\tilde K) z] =
    \sum_{i \neq j} [\log(\tilde K)]_{ij}^2 \leq
    \sum_{i=1}^n \log(1 + \lambda_j/\sigma^2)^2.
\]
If the eigenvalues of the kernel matrix without noise
decay rapidly enough compared to $\sigma$, the variance will
be small compared to the magnitude of
$\tr (\log \tilde{K}) =
 2 n \log \sigma + \sum_{i=1}^n \log(1 + \lambda_j/\sigma^2)$.
Hence, we need fewer probe vectors to obtain reasonable accuracy than
one would expect from bounds that are blind to the matrix structure.
In our experiments, we typically only use 5--10 probes --- and we use
the sample variance across these probes to estimate {\em a posteriori}
the stochastic component of the error in the log likelihood
computation.  If we are willing to estimate the Hessian of the log
likelihood, we can increase rates of convergence for finding
kernel hyperparameters.

The Chebyshev approximation scheme requires $O(\sqrt{\kappa}
\log(\kappa/\epsilon))$ steps to obtain an $O(\epsilon)$ approximation
error in computing $z^T \log(\tilde K) z$, where $\kappa = \lambda_{\max}/\lambda_{\min}$ is the condition
number of $\tilde{K}$~\cite{han2015large}.  This behavior is
independent of the distribution of eigenvalues within the interval
$[\lambda_{\min}, \lambda_{\max}]$, and is close to optimal when
eigenvalues are spread quasi-uniformly across the interval.
Nonetheless, when the condition number is large, convergence may be
quite slow.  The Lanczos approach converges at least twice as fast as
Chebyshev in general~\cite[Remark 1]{ubarufast}, and converges much
more rapidly when the eigenvalues are {\em not} uniform within the
interval, as is the case with log determinants of many kernel
matrices.  Hence, we recommend the Lanczos approach over the Chebyshev
approach in general.  In all of our experiments, the
error associated with approximating $z^T \log(\tilde K) z$ by Lanczos was
dominated by other sources of error.

\section{Experiments}
\label{sec:experiments}
We test our stochastic trace estimator with both Chebyshev and Lanczos
approximation schemes on:
(1) a sound time series with missing data, using a GP with an RBF kernel;
(2) a three-dimensional space-time precipitation data set with over half a million training points,
using a GP with an RBF kernel; (3)
a two-dimensional tree growth data set using a log-Gaussian Cox
process model with an RBF kernel;
(4) a three-dimensional space-time crime datasets with a log-Gaussian
Cox model with Mat\'ern 3/2 and spectral mixture kernels; and
(5) a high-dimensional
feature space using the deep kernel learning framework \citep{wilson2016deep}.
In the supplementary material we also include several additional experiments to
illustrate particular aspects of our approach, including kernel hyperparameter
recovery, diagonal
corrections (Section \ref{sup:diagcorrection}),
and surrogate methods (Section \ref{sec:rbf_interp}).  Throughout we use the
SKI method \citep{wilson2015kernel} of Eq.~\eqref{eqn: ski} for fast MVMs.
We find that the Lanczos and surrogate methods are able to do kernel
recovery and inference significantly faster and more accurately than
competing methods.

\subsection{Natural sound modeling}
\label{subsec:sound}
\begin{figure}[!ht]
    \centering
    \subfigure[\scriptsize Sound data]{\label{fig:sound_data}
    \includegraphics[width=0.23\textwidth,trim=0.4cm 0cm 3.5cm 1.0cm,clip]{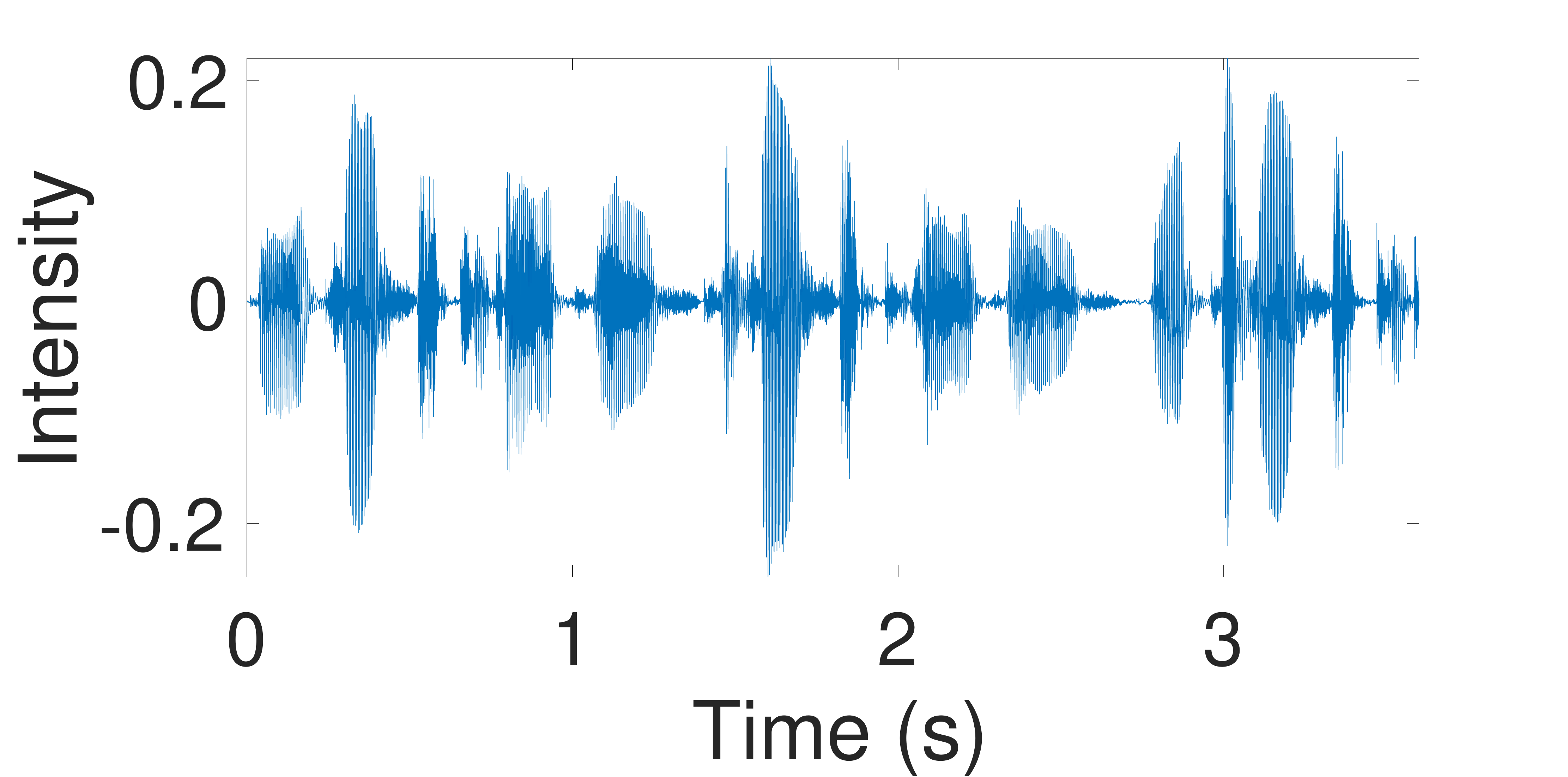}}
    \subfigure[\scriptsize Recovery time]{\label{fig:sound_recovery}
    \includegraphics[width=0.23\textwidth,trim=0.4cm 0cm 2.0cm 0cm,clip]{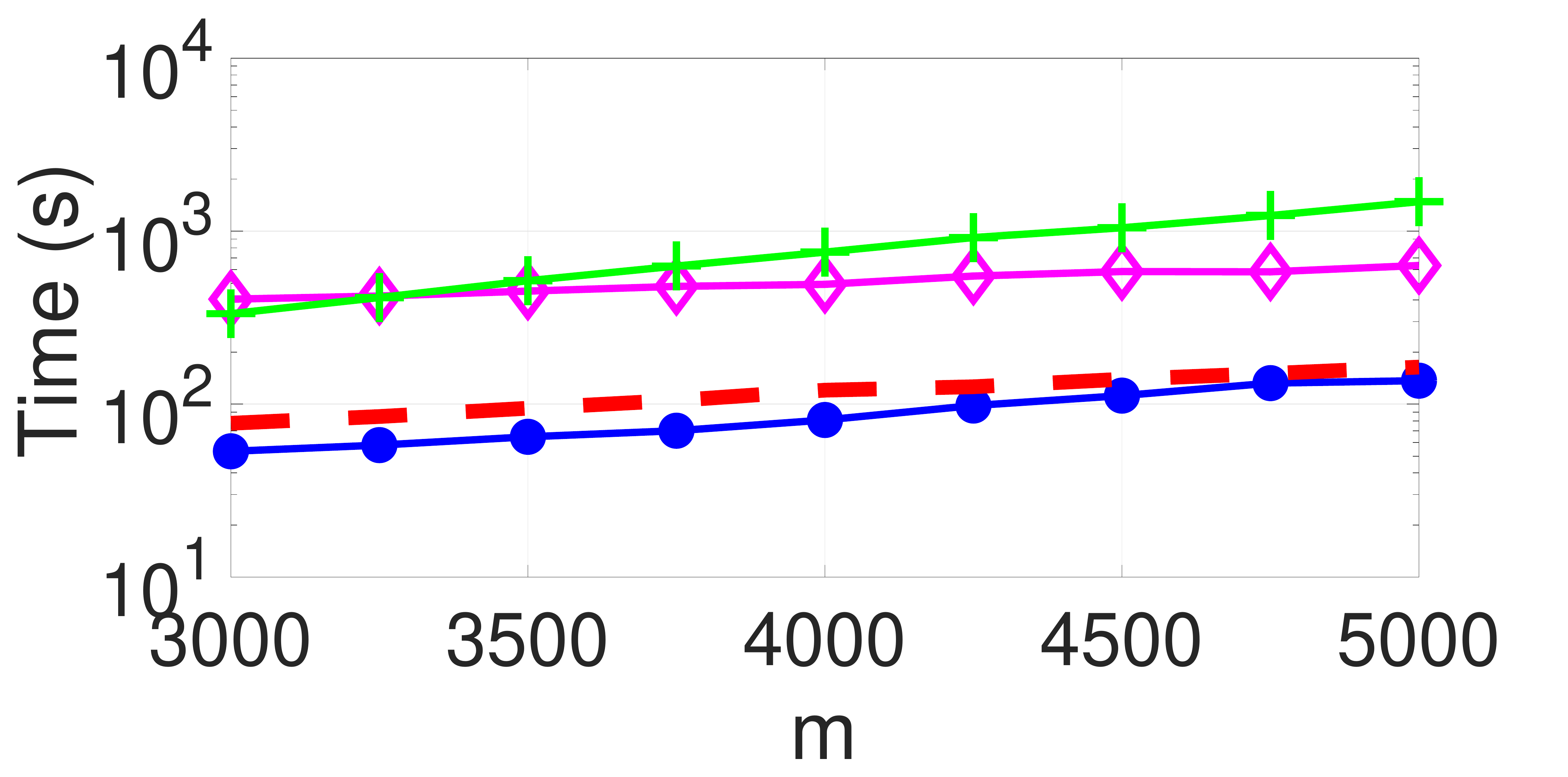}}
    \subfigure[\scriptsize Inference time]{\label{fig:sound_inference}
    \includegraphics[width=0.23\textwidth,trim=0.4cm 0cm 1.5cm 0.5cm,clip]{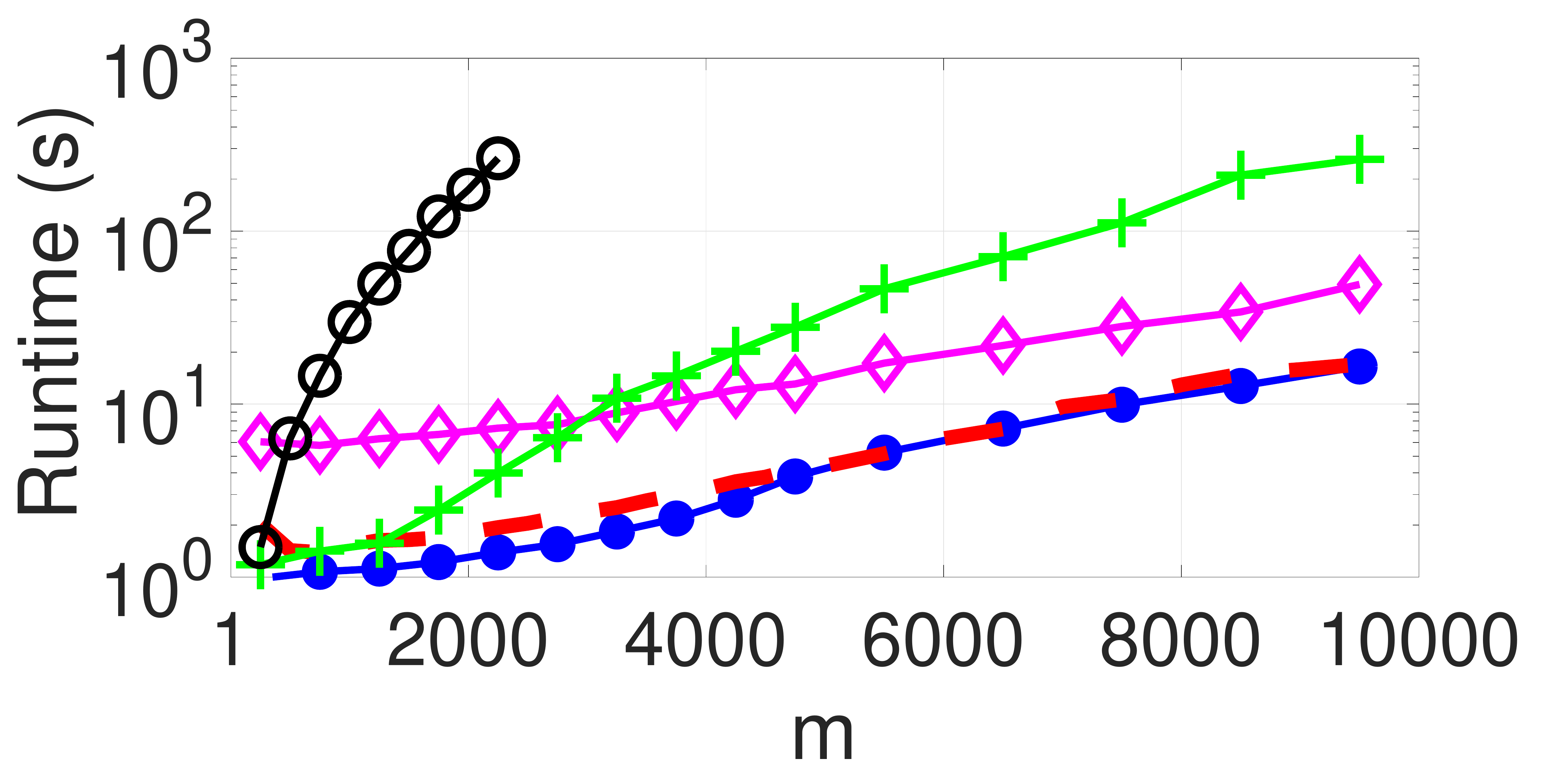}}
    \subfigure[\scriptsize SMAE]{\label{fig:sound_smae}
    \includegraphics[width=0.23\textwidth,trim=0.4cm 0cm 2.5cm 0.5cm,clip]{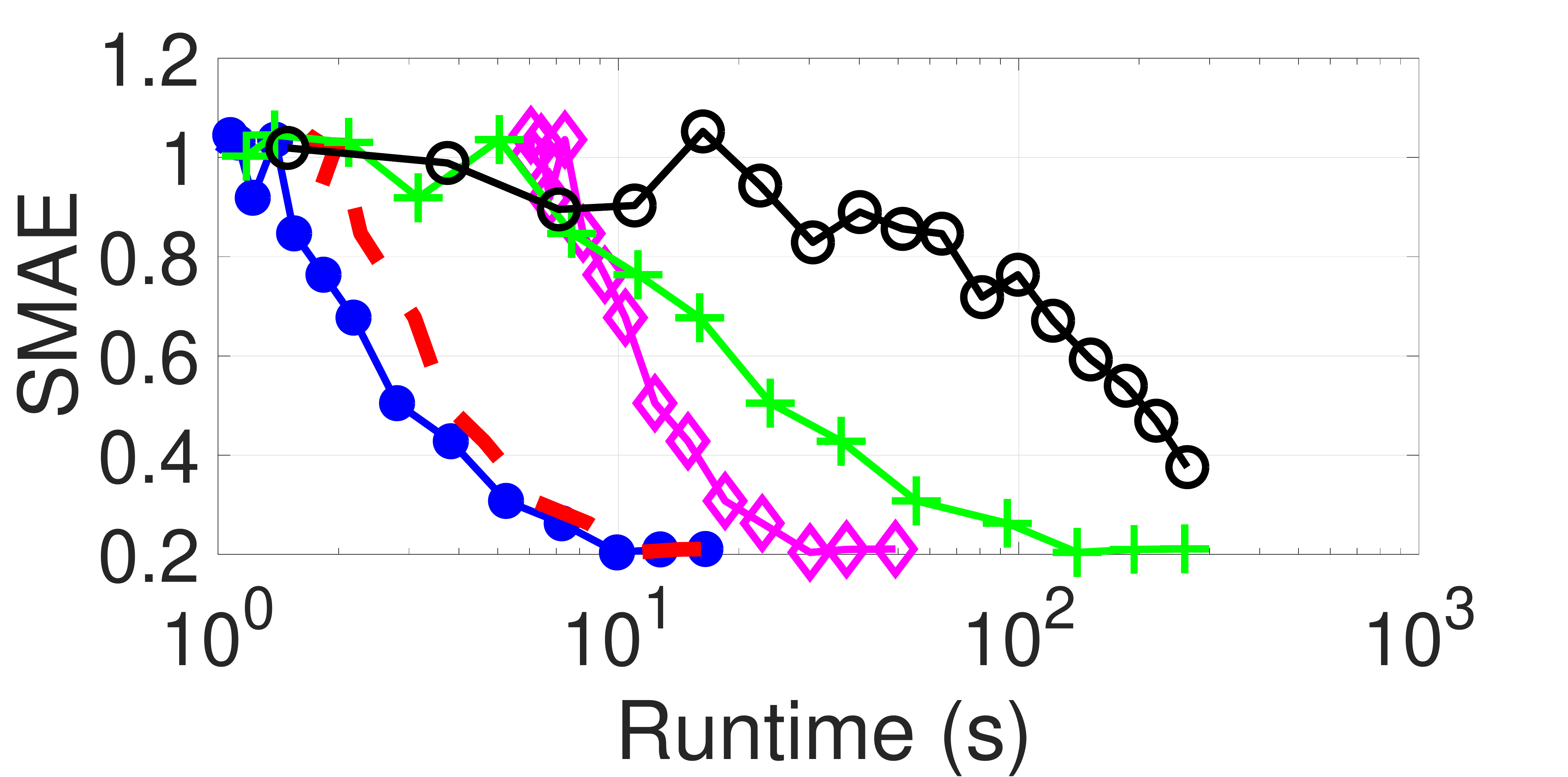}}
    \caption{Sound modeling using 59,306 training points and 691 test points. The
    intensity of the time series can be seen in (a). Train time for RBF kernel hyperparameters is in (b) and the time for inference is in (c). The standardized mean absolute error (SMAE) as a
    function of time for an evaluation of the marginal likelihood and all derivatives
    is shown in (d). Surrogate is ({\color{blue} ------}),
    Lanczos is ({\color{red} - - -}), Chebyshev is {(\color{magenta} --- $\diamond$ ---}),
    scaled eigenvalues is ({\color{green} --- + ---}),
    and FITC is ({\color{black} --- o ---}).}
    \label{fig:sound_modeling}
\end{figure}

Here we consider the natural sound benchmark in \cite{wilson2015kernel},
shown in Figure \ref{fig:sound_data}.
Our goal is to recover contiguous missing regions in a waveform with
$n = 59,306$ training points.  We
exploit Toeplitz structure in the $K_{UU}$ matrix
of our SKI approximate kernel for accelerated MVMs.

The experiment in \cite{wilson2015kernel} only considered scalable inference and prediction,
but not hyperparameter learning, since the scaled eigenvalue approach requires all
the eigenvalues for an $m \times m$ Toeplitz matrix, which can be computationally
prohibitive with cost $\mathcal{O}(m^2)$.  However, evaluating the marginal likelihood on
this training set is not an obstacle for Lanczos and Chebyshev since we can use fast
MVMs with the SKI approximation at a cost of
$\mathcal{O}(n + m \log m)$.

In Figure \ref{fig:sound_recovery}, we show how Lanczos, Chebyshev and
surrogate approaches scale with the number of inducing points $m$ compared to
the scaled eigenvalue method and FITC.  We use 5 probe vectors and
25 iterations for Lanczos, both when building the surrogate and for
hyperparameter learning with Lanczos. We also use 5 probe vectors for
Chebyshev and 100 moments. Figure \ref{fig:sound_recovery}
shows the runtime of the hyperparameter learning
phase for different numbers of inducing points $m$, where Lanczos
and the surrogate are clearly more efficient than scaled eigenvalues and Chebyshev.
For hyperparameter learning, FITC took several hours to run, compared to minutes
for the alternatives; we therefore exclude FITC from Figure~\ref{fig:sound_recovery}.
Figure \ref{fig:sound_inference} shows the time to do inference on the 691 test points,
while \ref{fig:sound_smae} shows the standardized mean absolute error (SMAE)
on the same test points. As expected, Lanczos and surrogate make
accurate predictions much faster than Chebyshev, scaled eigenvalues, and FITC.
In short, Lanczos and the surrogate approach are much faster than
alternatives for hyperparameter learning with a large number of inducing points
and training points.

\subsection{Daily precipitation prediction}
This experiment involves precipitation data from the year of 2010
collected from around $5500$ weather stations in the
US\footnote{\url{https://catalog.data.gov/dataset/u-s-hourly-precipitation-data}}. The
hourly precipitation data is preprocessed into daily data if full information
of the day is available. The dataset has $628,474$ entries in terms of
precipitation per day given the date, longitude and latitude. We randomly
select $100,000$ data points as test points and use the remaining
points for training. We then perform hyperparameter learning and prediction
with the RBF kernel, using Lanczos, scaled eigenvalues, and exact methods.

For Lanczos and scaled eigenvalues, we optimize the hyperparameters on
the subset of data for January 2010, with an induced grid of $100$ points
per spatial dimension and $300$ in the temporal dimension. Due to memory
constraints we only use a subset of $12,000$ entries for training with the exact method.
While scaled eigenvalues can perform well when fast eigendecompositions are possible,
as in this experiment, Lanczos nonetheless still runs faster and with slightly lower MSE.

\begin{table}[!ht]
    \centering
    \begin{tabular}{| c | c | c| c | c |}
        \hline
        Method  & $n$   & $m$ & MSE & Time [min]\\ \hline
        Lanczos & 528k  & 3M  & 0.613 &  14.3\\ \hline
        Scaled eigenvalues     & 528k  & 3M  & 0.621 &  15.9\\ \hline
        Exact   & 12k   & -   & 0.903 &  11.8\\ \hline
    \end{tabular}
    \caption{Prediction comparison for the daily precipitation data showing the
    number of training points $n$, number of induced grid points $m$, the mean squared
    error, and the inference time.}
    \label{tab:precip}
\end{table}
Incidentally, we are able to use 3 \emph{million} inducing points in Lanczos and scaled eigenvalues,
which is enabled by the SKI representation \citep{wilson2015kernel} of covariance matrices, for a
a very accurate approximation.  This number of inducing points $m$ is
unprecedented for typical alternatives which scale as $\mathcal{O}(m^3)$.

\subsection{Hickory data}
In this experiment, we apply Lanczos to the log-Gaussian Cox process
model with a Laplace approximation for the posterior
distribution. We use the RBF kernel and
the Poisson likelihood in our model.  The scaled eigenvalue method
does not apply directly to non-Gaussian likelihoods; we thus applied
the scaled eigenvalue method in \citep{wilson2015kernel} in conjunction
with the Fiedler bound in \citep{flaxman2015fast} for the scaled
eigenvalue comparison.  Indeed, a key advantage of the Lanczos
approach is that it can be applied whenever fast MVMs are available,
which means no additional approximations such as the Fiedler bound
are required for non-Gaussian likelihoods.

\begin{figure}[!ht]
    \centering
    \subfigure[\scriptsize Point pattern data]{\label{fig:hpoint}
      \includegraphics[width=0.18\textwidth,height=0.16\textwidth,
                       trim = 1cm 1cm 1cm 1cm,clip]{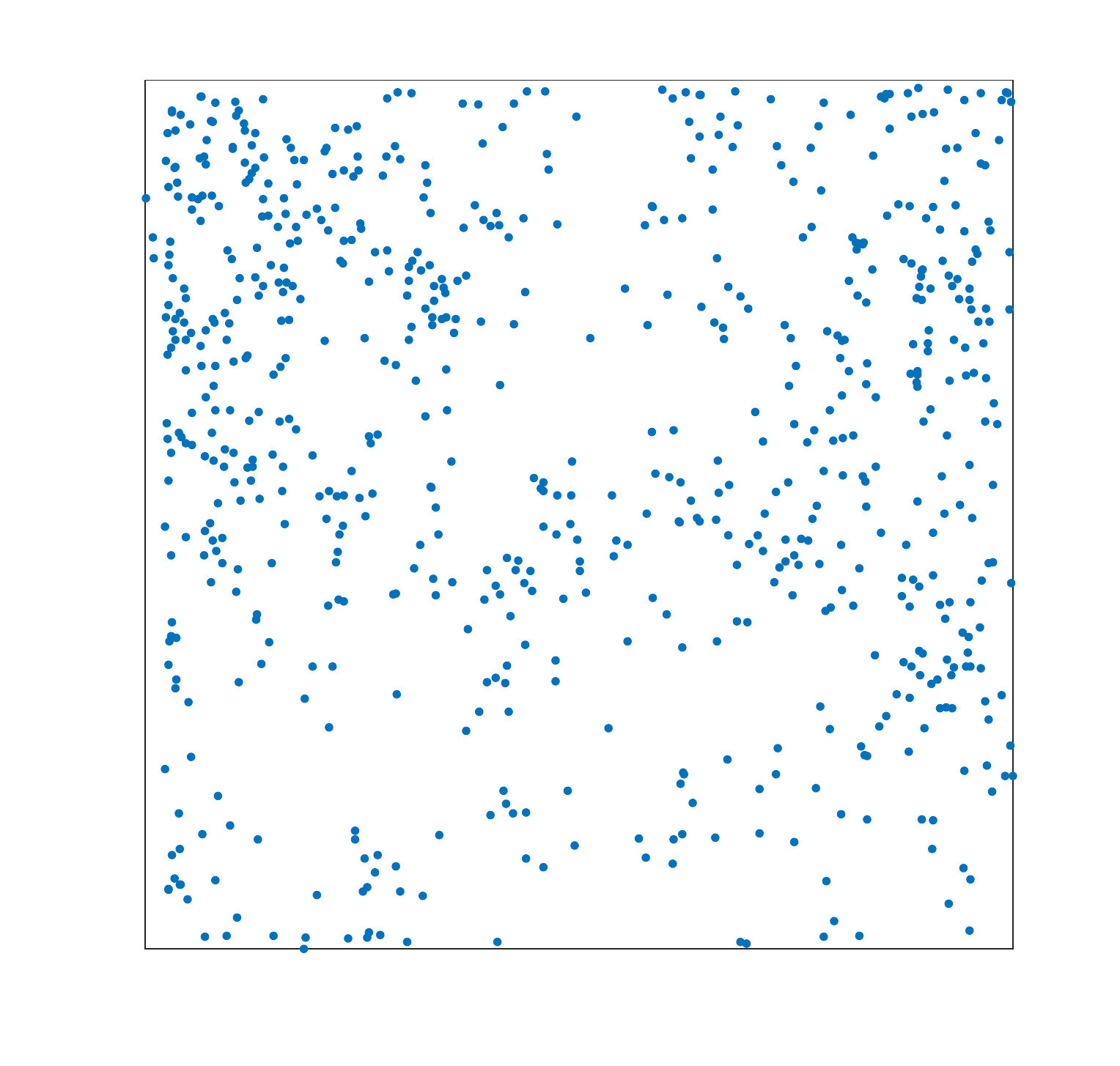}}
    \subfigure[\scriptsize Prediction by exact]{\label{fig:hickory_exact}
      \includegraphics[width=0.18\textwidth,height=0.16\textwidth,
                       trim = 1cm 1cm 1cm 1cm,clip]{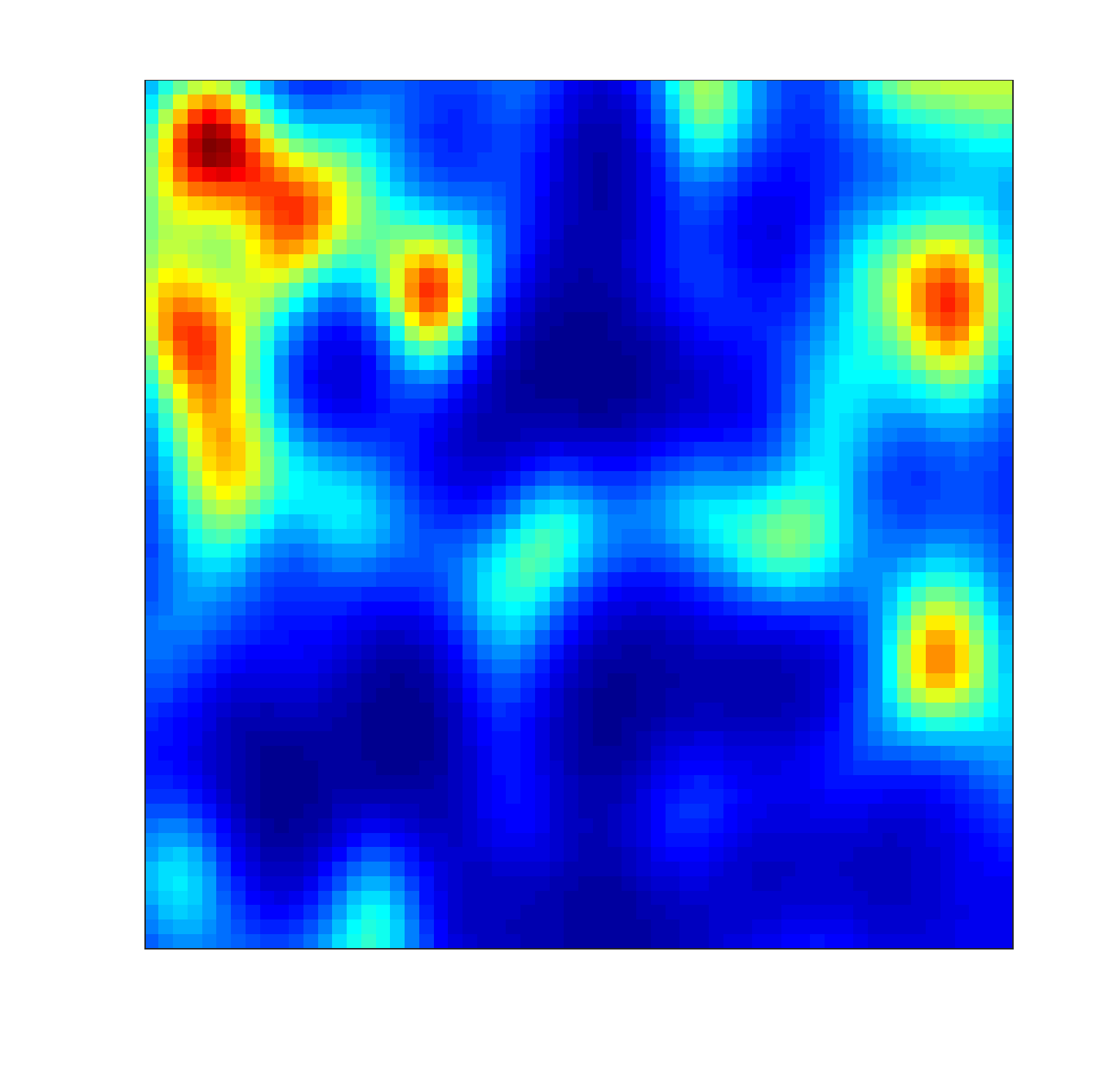}}
    \subfigure[\scriptsize Scaled eigenvalues]{\label{fig:hickory_ski}
      \includegraphics[width=0.18\textwidth,height=0.16\textwidth,
                       trim = 1cm 1cm 1cm 1cm,clip]{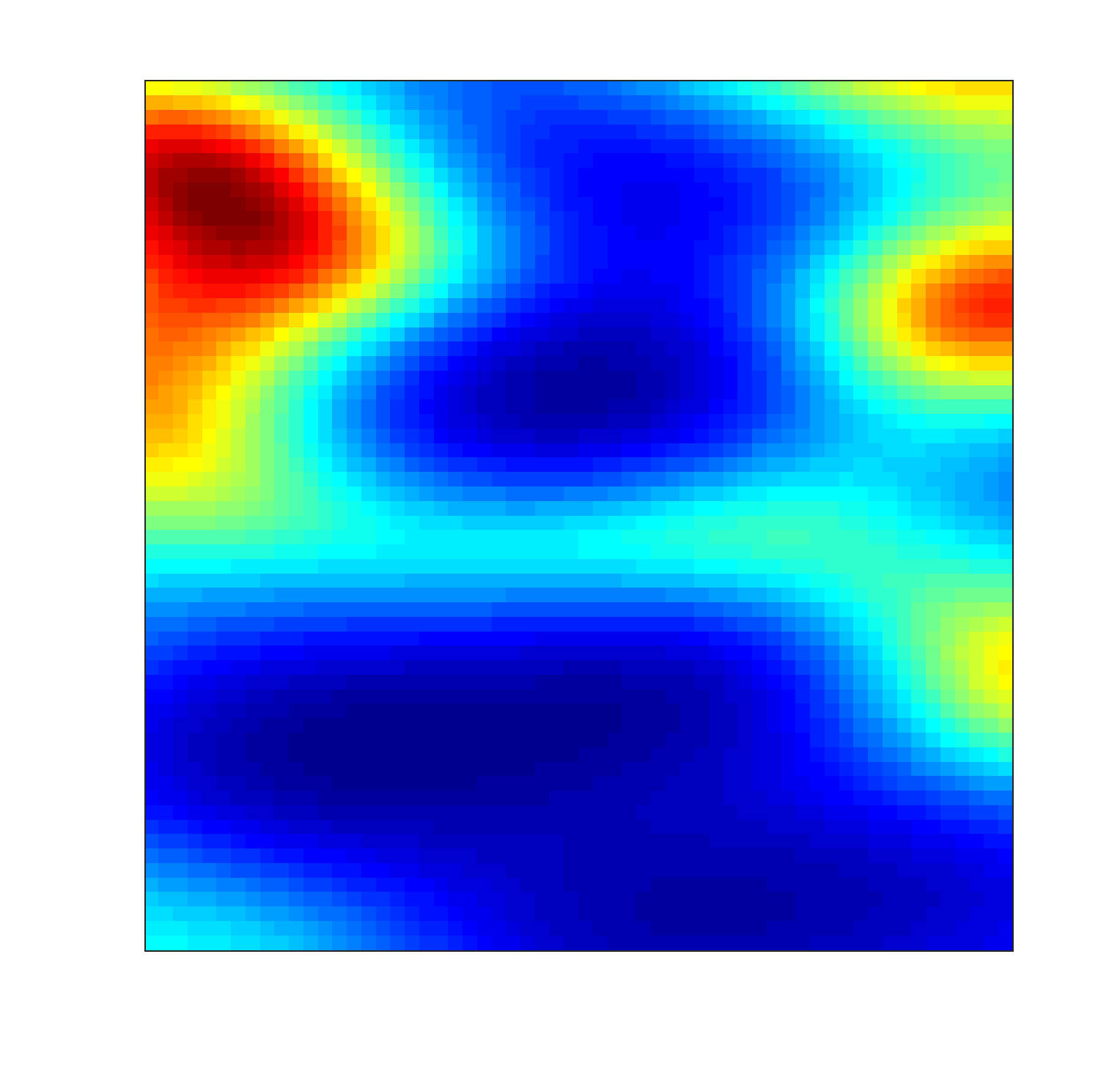}}
    \subfigure[\scriptsize Lanczos]{\label{fig:hickory_lan}
      \includegraphics[width=0.18\textwidth,height=0.16\textwidth,
                       trim = 1cm 1cm 1cm 1cm,clip]{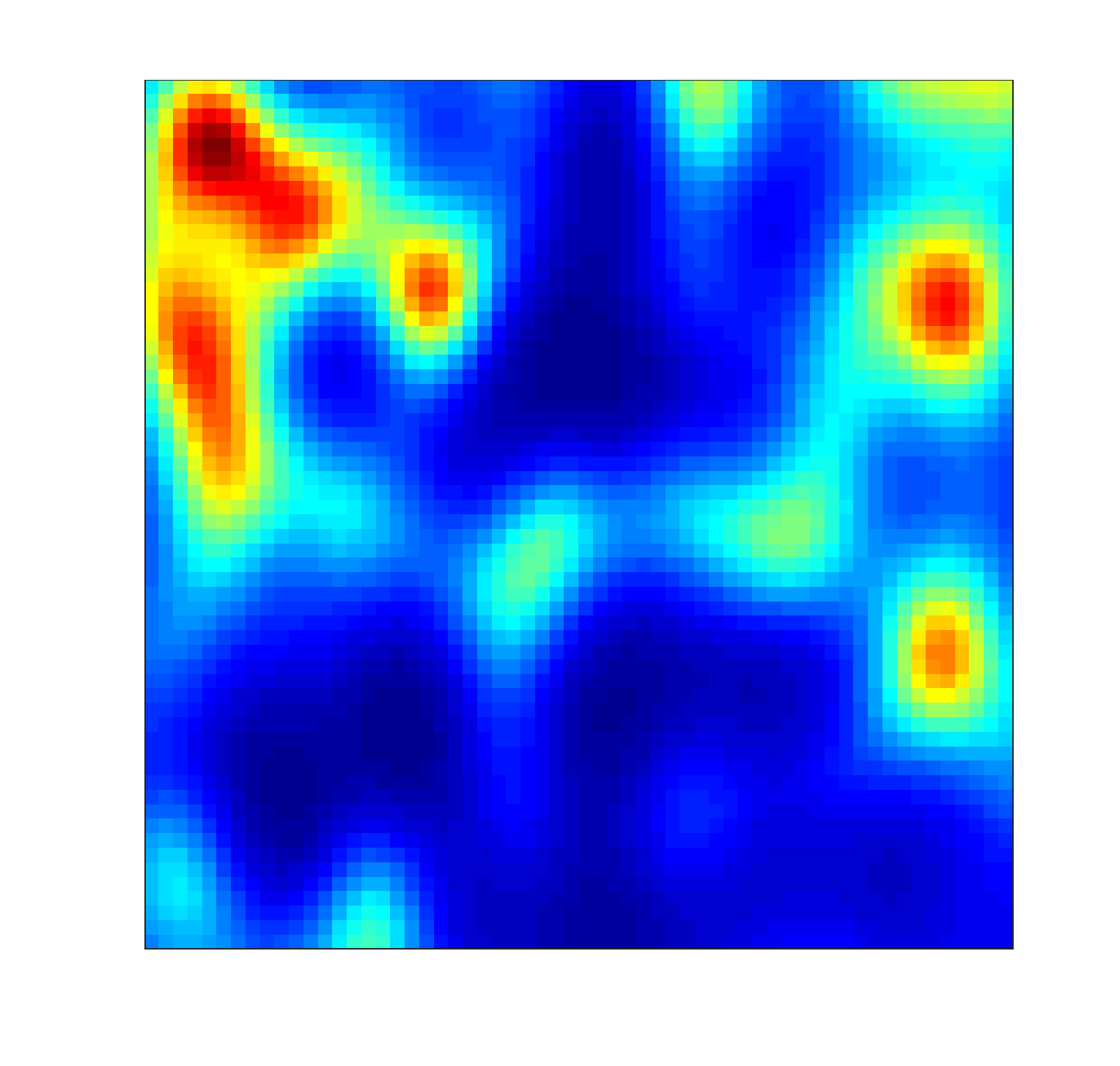}}
    \caption{Predictions by exact, scaled eigenvalues, and Lanczos on the Hickory dataset.}
    \label{fig:hickory}
\end{figure}

This dataset, which comes from the R package
{\tt spatstat}, is a point pattern of $703$ hickory trees in a forest in
Michigan. We discretize the area into a $60 \times 60$ grid and fit
our model with exact, scaled eigenvalues, and Lanczos. We see in Table
\ref{tab:hickory} that Lanczos recovers hyperparameters that are much
closer to the exact values than the scaled eigenvalue approach.
Figure \ref{fig:hickory} shows that the predictions by Lanczos
are also indistinguishable from the exact computation.

\begin{table}[!ht]
    \centering
    \begin{tabular}{| c | c | c| c | c | c|}
        \hline
        Method & $s_f$ & $\ell_1$ & $\ell_2$ & $-\log p(y|\theta)$ & Time [s]\\
        \hline
        Exact & $0.696$  & $0.063$  & $0.085$ &  $1827.56$ & $465.9$\\
        \hline
        Lanczos & $0.693$   & $0.066$   & $0.096$ &  $1828.07$ & $21.4$\\
        \hline
        Scaled eigenvalues & $0.543$  & $0.237$  & $0.112$ &  $1851.69$ & $2.5$\\
        \hline
    \end{tabular}
    \caption{Hyperparameters recovered on the Hickory dataset.}
    \label{tab:hickory}
\end{table}

\vspace{-5mm}
\vspace{-2mm}

\subsection{Crime prediction}
In this experiment, we apply Lanczos with the spectral mixture kernel to
the crime forecasting problem considered in \cite{flaxman2015fast}.
This dataset consists of $233,088$
incidents of assault in Chicago from January 1, 2004 to December 31,
2013. We use the first $8$ years for training and attempt to predict
the crime rate for the last $2$ years. For the spatial dimensions, we
use the log-Gaussian Cox process model, with the Mat\'ern-5/2
kernel, the negative binomial likelihood, and the Laplace approximation for
the posterior. We use a spectral mixture kernel with $20$
components and an extra constant component for the temporal
dimension. We discretize the data into a $17 \times 26$ spatial grid
corresponding to $1$-by-$1$ mile grid cells. In the temporal
dimension we sum our data by weeks for a total of $522$ weeks.
After removing the cells that are outside Chicago, we
have a total of $157,644$ observations.

The results for Lanczos and scaled eigenvalues (in conjunction with
the Fiedler bound due to the non-Gaussian likelihood) can be seen in Table
\ref{tab:chicago_homicide}. The Lanczos method used $5$ Hutchinson probe
vectors and $30$ Lanczos steps. For both methods we allow $100$ iterations of
LBFGS to recover hyperparameters and we often observe early convergence.
While the RMSE for Lanczos and scaled eigenvalues happen to be close
on this example, the
recovered hyperparameters using scaled eigenvalues are very
different than for Lanczos.  For example, the scaled eigenvalue method learns a much
larger $\sigma^2$ than Lanczos, indicating
model misspecification.  In general, as the data become increasingly non-Gaussian
the Fiedler bound (used for fast scaled eigenvalues on non-Gaussian likelihoods) will
become increasingly misspecified, while Lanczos will be unaffected.

\begin{table}[!h]
    \centering
    \scalebox{0.9}{
        \begin{tabular}{| c | c | c | c | c | c| c| c|}
            \hline
            Method & $\ell_1$ & $\ell_2$ & $\sigma^2$ &
            $\text{T}_{\text{recovery}}$[s] & $\text{T}_{\text{prediction}}$[s]
            & $\text{RMSE}_{\text{train}}$ & $\text{RMSE}_{\text{test}}$ \\
            \hline
            Lanczos & $0.65$ & $0.67$ & $69.72$ & $264$ & $10.30$
            & $1.17$ & $1.33$\\
            \hline
            Scaled eigenvalues & $0.32$ & $0.10$ & $191.17$ & $67$ & $3.75$
            & $1.19$  & $1.36$ \\
            \hline
        \end{tabular}
    }
    \caption{Hyperparameters recovered, recovery time and RMSE for Lanczos and
           scaled eigenvalues on the Chicago assault data. Here $\ell_1$ and
           $\ell_2$ are
           the length scales in spatial dimensions and $\sigma^2$ is the noise
           level. $\text{T}_{\text{recovery}}$ is the time for recovering
           hyperparameters. $\text{T}_{\text{prediction}}$ is the time for
           prediction at all $157,644$ observations (including training and
           testing).}
           \label{tab:chicago_homicide}
\end{table}

\vspace{-4mm}
\subsection{Deep kernel learning}
To handle high-dimensional datasets, we bring our methods
into the deep kernel learning framework \cite{wilson2016deep} by replacing the final layer
of a pre-trained deep neural network (DNN) with a GP.
This experiment uses the gas sensor dataset from the UCI machine
learning repository. It has $2565$ instances with $128$ dimensions.
We pre-train a DNN, then
attach a Gaussian process with RBF kernels to the two-dimensional
output of the second-to-last layer. We then further train all parameters
of the resulting kernel, \emph{including} the weights of the DNN,
through the GP marginal likelihood.
In this example, Lanczos and the scaled eigenvalue approach perform similarly
well.  Nonetheless, we see that Lanczos can effectively be
used with SKI on a high dimensional problem to train hundreds of thousands
of kernel parameters.

\begin{table}[!h]
    \centering
    \scalebox{0.9}{
        \begin{tabular}{| c | c | c | c |}
            \hline
            Method & DNN & Lanczos & Scaled eigenvalues \\ \hline
            RMSE & $0.1366\pm 0.0387$ & $0.1053\pm0.0248$ & $0.1045\pm 0.0228$\\ \hline
            Time [s]& $0.4438$ & $2.0680$ & $1.6320$\\ \hline
        \end{tabular}
    }
    \caption{Prediction RMSE and per training iteration runtime.}
    \label{tab:dkl}
\end{table}
\vspace{-4mm}

\section{Discussion}
\label{sec:discussion}
There are many cases in which fast MVMs can be achieved,
but it is difficult or impossible to efficiently compute a log
determinant.  We have developed a framework for scalable and
accurate estimates of a log determinant and its derivatives relying
only on MVMs.  We particularly consider scalable kernel learning,
showing the promise of stochastic Lanczos estimation combined with
a pre-computed surrogate model.  We have shown the scalability and
flexibility of our approach through experiments with kernel learning for
several real-world data sets using both Gaussian and non-Gaussian
likelihoods, and highly parametrized deep kernels.

Iterative MVM approaches have great promise for future exploration.
We have only begun to explore their significant generality.  In addition
to log determinants, the methods presented here could be adapted
to fast posterior sampling, diagonal estimation, matrix square roots,
and many other standard operations.  The proposed methods only depend
on fast MVMs---and the structure necessary for fast MVMs often exists,
or can be readily created.  We have here
made use of SKI \citep{wilson2015kernel} to create such structure.
But other approaches, such as stochastic variational methods
\citep{hensman2013uai}, could be used or combined
with SKI for fast MVMs, as in \citep{wilson2016stochastic}.  Moreover, iterative MVM methods naturally
harmonize with GPU acceleration, and are therefore likely to increase
in their future applicability and popularity.  Finally, one could explore
the ideas presented here for scalable higher order derivatives,
making use of Hessian methods for greater convergence rates.

\bibliography{references}
\bibliographystyle{unsrtnat}

\appendix
\label{sec:supp}
\section{Background}
\label{sup:background}

Two popular covariance kernels are the RBF kernel
\[
    k_{\text{RBF}}(x, x') = s_f^2 \exp\left(\frac{\|x-x'\|^2}{2\ell^2}\right) \\
\]
and the Mat\'ern kernel
\begin{equation*}
    \resizebox{0.51\textwidth}{!}{%
        $k_{\text{Mat},{\nu}}(x,x') =  s_f^2\frac{2^{1-\nu}}{\Gamma(\nu)}
        \left(\sqrt{2\nu}\frac{\|x-x'\|}{\ell}\right)^\nu
        K_\nu\left(\sqrt{2\nu}\frac{\|x-x'\|}{\ell}\right)$
    }
\end{equation*}
where $1/2$, $3/2$, and $5/2$ are popular choices for $\nu$ to model
heavy-tailed correlations between function values.  The spectral
behavior of these and other kernels has been well-studied for years,
and we recommend~\citep{wathen2015spectral} for recent results.
Particularly relevant to our discussion is a theorem due to Weyl,
which says that if a symmetric kernel has $\nu$ continuous
derivatives, then the eigenvalues of the associated integral operator
decay like
$|\lambda_n| = o(n^{-\nu-1/2})$.  Hence, the eigenvalues of
kernel matrices for the smooth
RBF kernel (and of any given covariance matrix based on that kernel)
tend to decay much more rapidly than those of the less
smooth Mat\'ern kernel, which has two derivatives at zero for $\nu =
5/2$, one derivative at zero for $\nu = 3/2$, and no derivatives at zero for
$\nu = 1/2$.  This matters to the relative performance of Chebyshev
and Lanczos approximations of the log determinant for large values of
$s_f$ and small values of $\sigma$ on the exact and approximate RBF
kernel.

\section{Methods}
\label{sup:methods}

\subsection{Scaled eigenvalue method}
\label{sup:scaledeigenvalues}
The scaled eigenvalue method was introduced in ~\citep{wilson2014fast}
to estimate $\log |K_{XX}+\sigma^2I|$, where $X$ consists of $n$ points.
The eigenvalues $\{\lambda_i\}_{i=1}^n$
of $K_{XX}$ can be approximated using the $n$ largest eigenvalues of a
covariance matrix $\tilde{K}_{YY}$ on a full grid with $m$
points such that $X \subset Y$. Specifically,
\[
    \log |K_{XX}+\sigma^2I| =
    \sum_{i=1}^n \log(\lambda_i + \sigma^2)
    \approx \sum_{i=1}^n \log\left(\frac{n}{m}\tilde{\lambda}_i + \sigma^2\right)
\]
The induced kernel $K_{UU}$ plays the role of $\tilde{K}_{YY}$ when the scaled
eigenvalue method is applied to SKI and the eigenvalues of $K_{UU}$ can
be efficiently computed. Assuming that the eigenvalues can be computed
efficiently is a much stronger assumption than our fast MVM based approach.

\subsection{Radial basis function surrogates}
\label{sup:surrogate}

Radial basis function (RBF) interpolation is one of the most popular approaches
to approximating scattered data in a general number of dimensions
\citep{buhmann2000radial,fasshauer2007meshfree,schaback2006kernel,wendland2004scattered}.
Given
distinct interpolation points
$\Theta=\{\theta^i\}_{i=1}^n$, the RBF model
takes the form
\begin{equation}
    \label{eq:rbf}
    s_{\Theta}(\theta) = \sum_{i=1}^n \lambda_i \varphi(\|x - \theta^i\|) + p(x)
\end{equation}
where the kernel $\varphi : \mathbb{R}_{\geq 0} \to \mathbb{R}$ is a
one-dimensional function and $p \in \Pi_{m-1}^d$, the space of polynomials with
$d$ variables of degree no more than $m-1$.
There are many possible choices for
$\varphi$ such as the cubic kernel $\varphi(r)=r^3$ and the thin-plate spline
kernel $\varphi(r)=r^2\,\log(r)$.
The coefficients $\lambda_i$ are determined
by imposing the interpolation conditions
$s_{\Theta}(\theta^i) = \log |K(\theta^i)|$ for
$i=1,\ldots,n$ and the discrete orthogonality condition
\begin{equation}
    \label{eq:disc_orthg}
    \sum_{i=1}^n \lambda_i q(\theta^i) = 0, \qquad \forall q \in \Pi_{m-1}^d.
\end{equation}
For appropriate RBF kernels, this linear system is nonsingular
provided that polynomials in $\Pi_{m-1}^d$ are uniquely determined by
their values on the interpolation set.

\subsection{Comparison to a reference kernel}
\label{sup:refkernel}

Suppose more generally that
$\tilde K = K + \sigma^2 I$ is an approximation to a reference
kernel matrix $\tilde K^{\mathrm{ref}} = K^{\mathrm{ref}} + \sigma^2 I$,
and let $E = K^{\mathrm{ref}}-K$.  Let $\mathcal{L}(\theta | y)$
and $\mathcal{L}^{\mathrm{ref}}(\theta | y)$ be the log likelihood
functions for the two kernels; then
\begin{align*}
    \mathcal{L}^{\mathrm{ref}}(\theta | y) &=
    \mathcal{L}(\theta | y) -
    \frac{1}{2} \left[
        \tr(\tilde K^{-1} E) -
        \alpha^T E \alpha
    \right] + O(\|E\|^2) \\
    \frac{\partial}{\partial \theta_i} \mathcal{L}^{\mathrm{ref}}(\theta | y) &=
    \frac{\partial}{\partial \theta_i} \mathcal{L}(\theta | y)
    - \frac{1}{2} \left[
      \operatorname{tr}\left(
        \tilde K^{-1} \frac{\partial E}{\partial \theta_i} -
        \tilde K^{-1} \frac{\partial \tilde K}{\partial \theta_i}
        \tilde K^{-1} E
      \right) -
      \alpha^T \frac{\partial E}{\partial \theta_i}  \alpha
\right] + O(\|E\|^2).
\end{align*}
If we are willing to pay the price of a few MVMs with $E$, we can use
these expressions to improve our maximum likelihood estimate.
Let $z$ and $w$ be independent probe vectors with
$g = \tilde K^{-1} z$ and $\hat{g} = \tilde K^{-1} w$.
To estimate the trace in the derivative computation, we
use the standard stochastic trace estimation approach
together with the observation that $\mathbb{E}[ww^T] = I$:
\[
   \operatorname{tr}\left(
     \tilde K^{-1} \frac{\partial E}{\partial \theta_i} -
     \tilde K^{-1} \frac{\partial \tilde K}{\partial \theta_i}
     \tilde K^{-1} E
   \right) =
   \mathbb{E}\left[
     g^T \frac{\partial E}{\partial \theta_i} z -
     g^T \frac{\partial K}{\partial \theta_i} w \hat{g}^T E z
   \right]
\]
This linearization may be used directly (with a stochastic estimator);
alternately, if we have an estimates for $\|E\|$ and $\|\partial
E/\partial \theta_i\|$, we can substitute these in order to get
estimated bounds on the magnitude of the derivatives.
Coupled with a similar estimator for the Hessian of the likelihood
function (described in the supplementary materials), we can use this
method to compute the maximum likelihood parameters for the fast
kernel, then compute a correction $-H^{-1} \nabla_{\theta} \mathcal{L}^{\mathrm{ref}}$
to estimate the maximum likelihood parameters of the reference kernel.

\section{Additional experiments}
\label{sup:experiments}

This section contains several experiments with synthetic data sets to illustrate
particular aspects of the method.

\subsection{1D cross-section plots}
\label{sup:1dcross}

\begin{figure}[!ht]
    \centering
    \subfigure[\scriptsize log marginal likelihood for the RBF kernel]{\label{fig:1dpert_a}
      \includegraphics[width=0.46\textwidth,trim=3.5cm 0cm 3.5cm 0cm,clip]{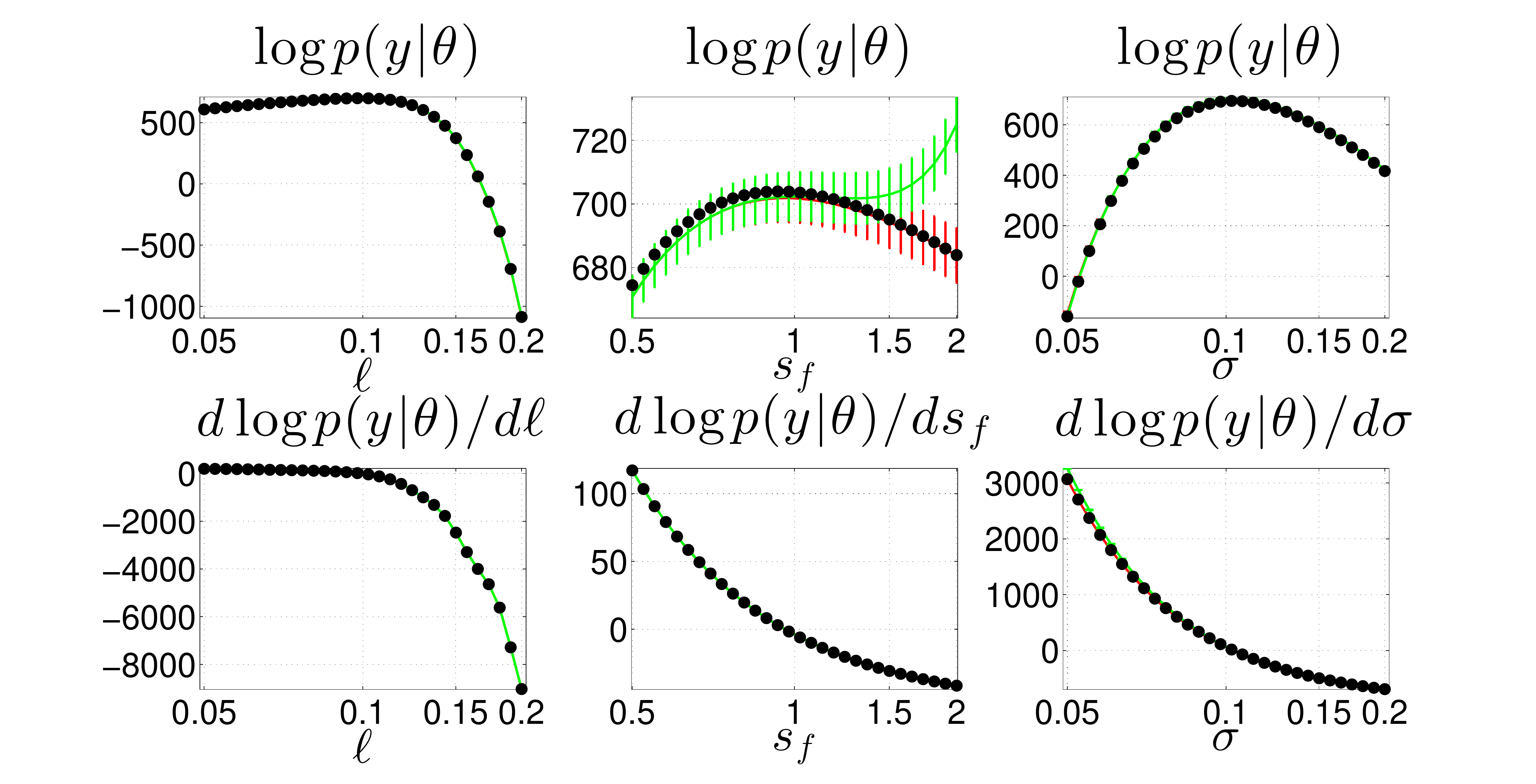}}
    \subfigure[\scriptsize log marginal likelihood for the Mat\'ern kernel]{\label{fig:1dpert_b}
      \includegraphics[width=0.46\textwidth,trim=3.5cm 0cm 3.5cm 0cm,clip]{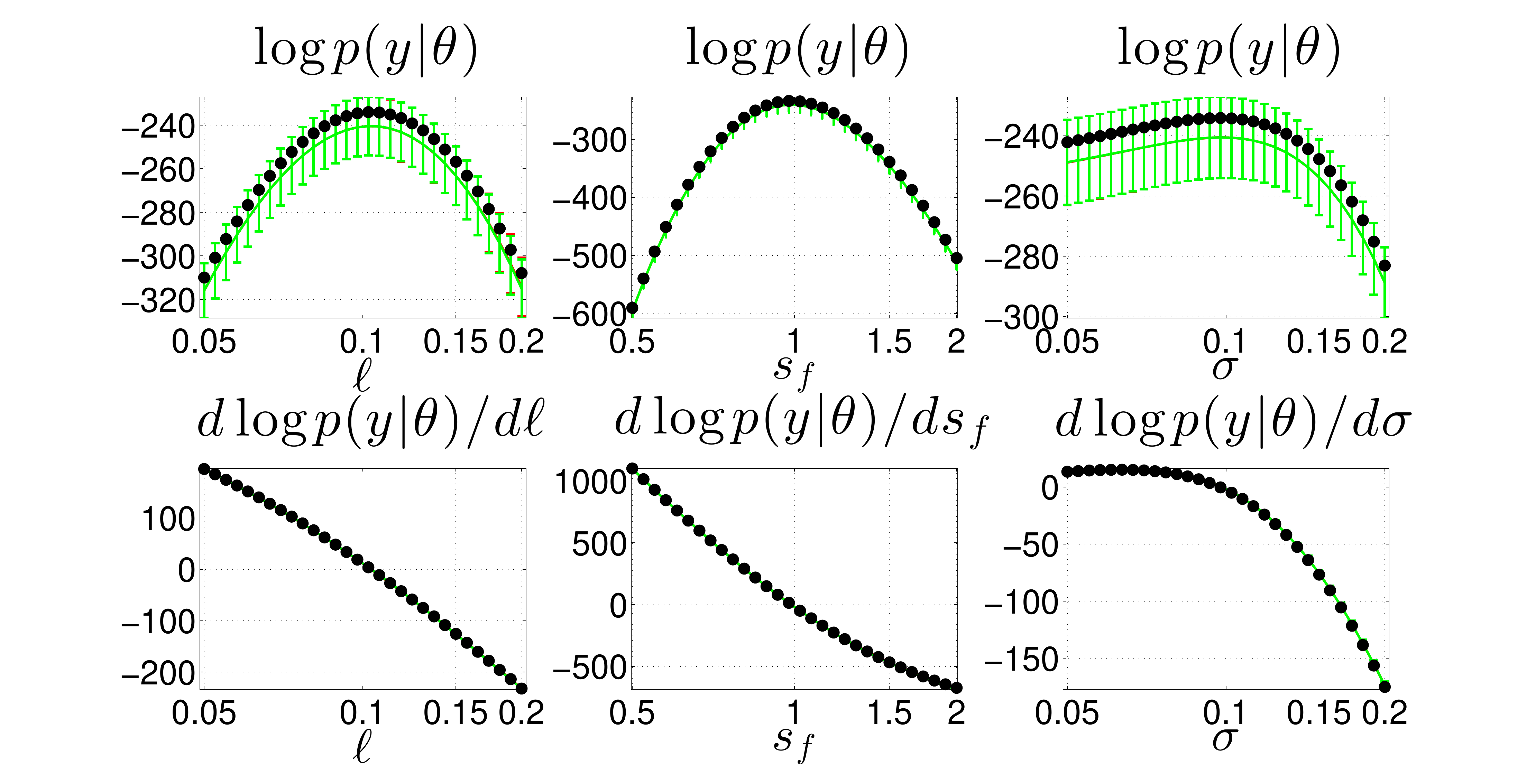}}
    \subfigure[\scriptsize log determinant for the RBF kernel]{\label{fig:1dpert_c}
      \includegraphics[width=0.46\textwidth,trim=3.5cm 0cm 3.5cm 0cm,clip]{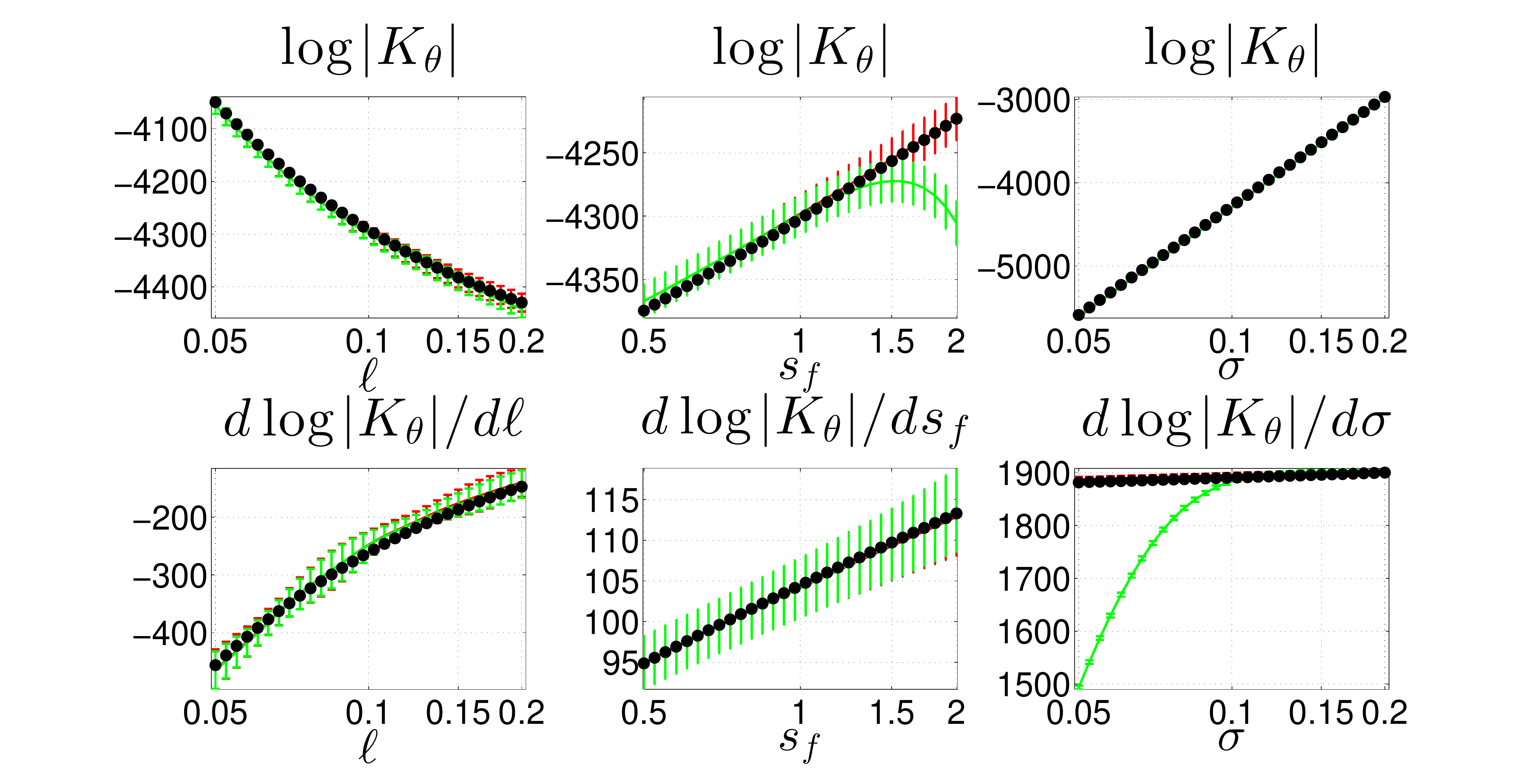}}
    \subfigure[\scriptsize log determinant for the Mat\'ern kernel]{\label{fig:1dpert_d}
      \includegraphics[width=0.46\textwidth,trim=3.5cm 0cm 3.5cm 0cm,clip]{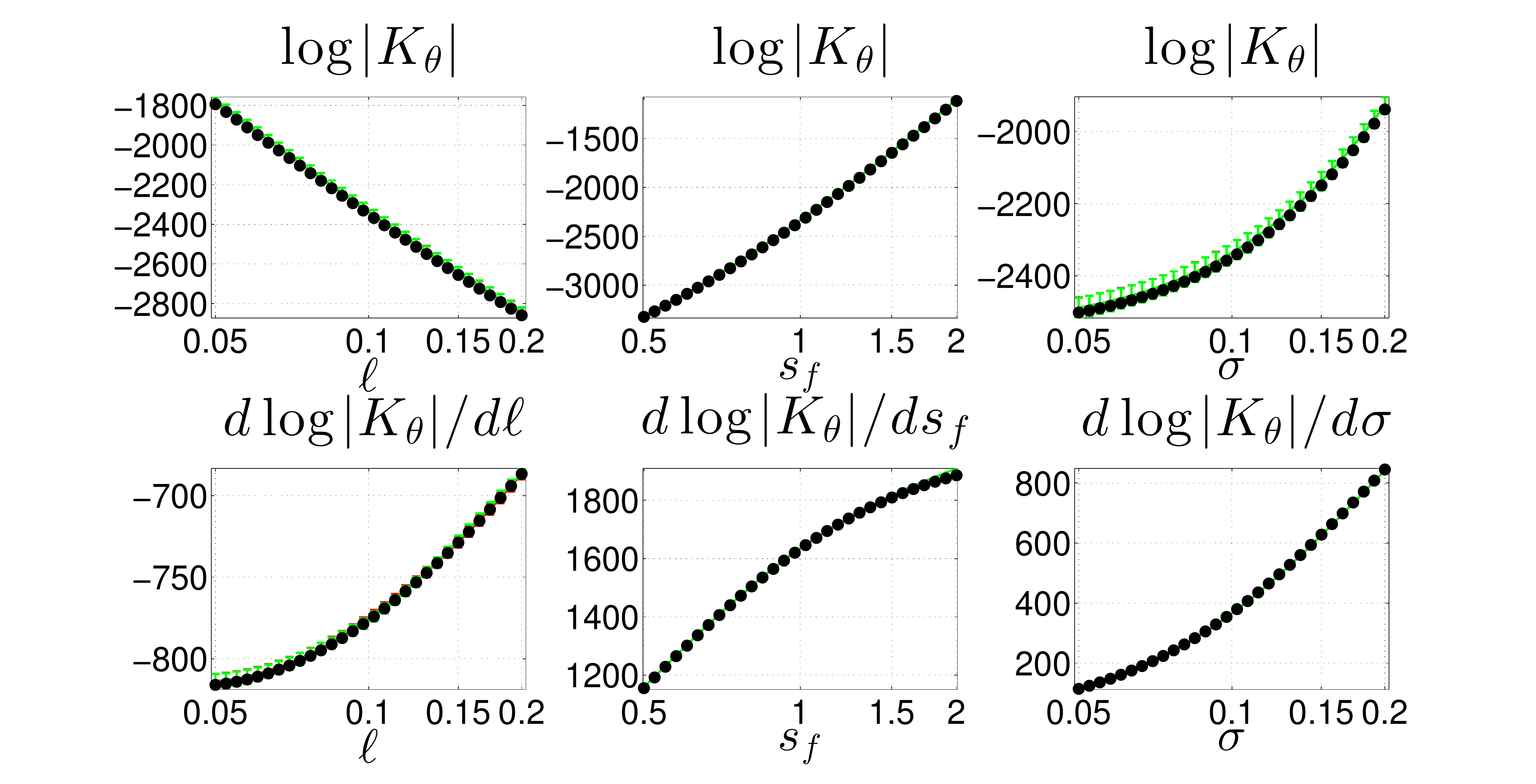}}
    \caption{1-dimensional perturbations for the exact RBF and Mat\'ern 1/2 kernel where the data
             is $1000$ equally spaced points in the interval $[0,4]$. The exact values
             are ($\bullet$), Lanczos is ({\color{red}-----}),
             Chebyshev is ({\color{green}-----}). The error bars of Lanczos and
             Chebyshev are $1$ standard deviation and were computed from $10$ runs with
             different probe vectors}
     \label{fig:1dpert}
\end{figure}

\begin{figure}[!ht]
    \centering
    \subfigure[\scriptsize log marginal likelihood for the RBF kernel]{\label{fig:1dpert_kissgp_a}
      \includegraphics[width=0.46\textwidth,trim=2.5cm 0cm 3.5cm 0cm,clip]{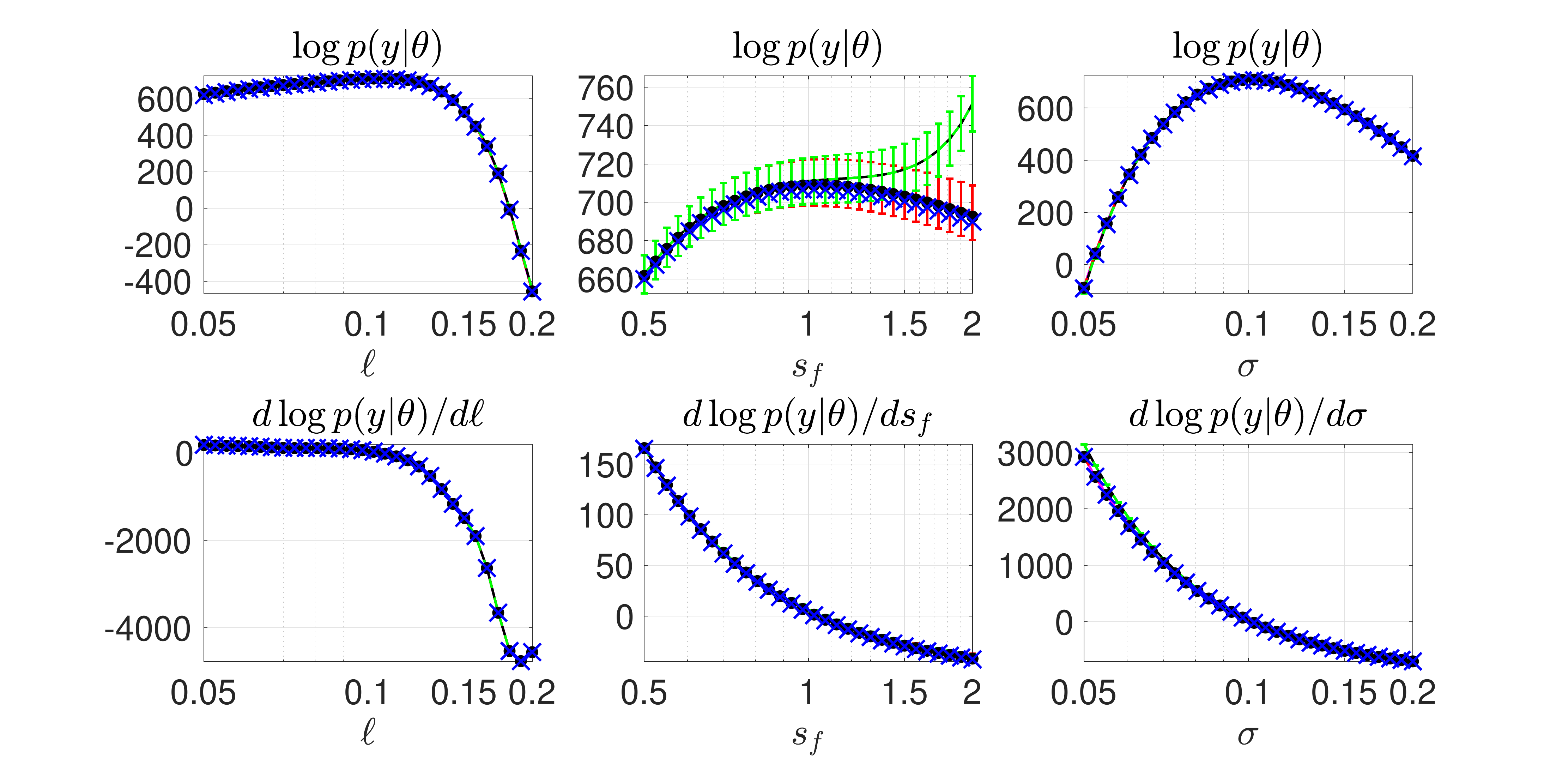}}
    \subfigure[\scriptsize log marginal likelihood for the Mat\'ern kernel]{\label{fig:1dpert_kissgp_b}
      \includegraphics[width=0.46\textwidth,trim=2.5cm 0cm 3.5cm 0cm,clip]{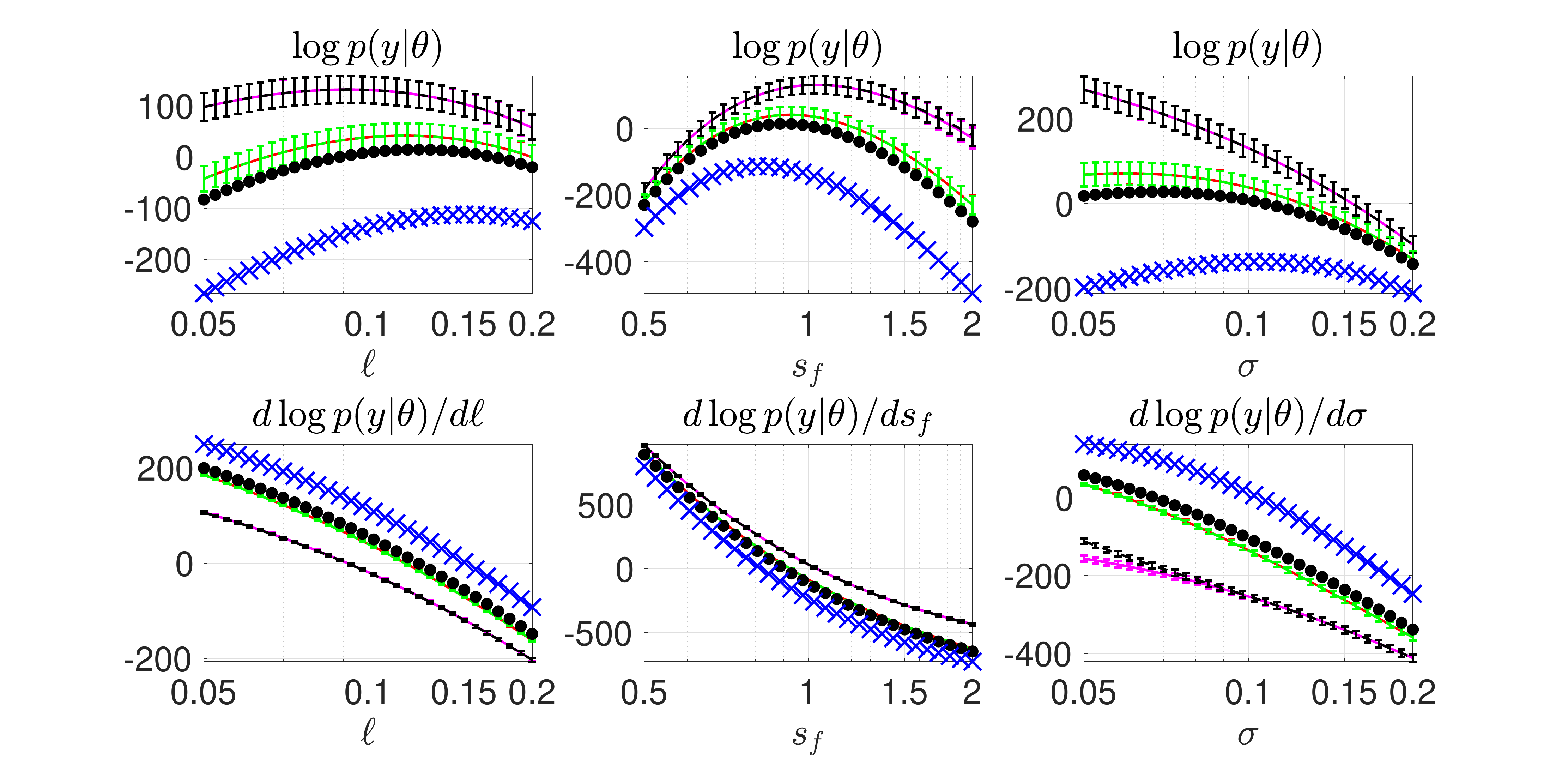}}
    \subfigure[\scriptsize log determinant for the RBF kernel]{\label{fig:1dpert_kissgp_c}
      \includegraphics[width=0.46\textwidth,trim=2.5cm 0cm 3.5cm 0cm,clip]{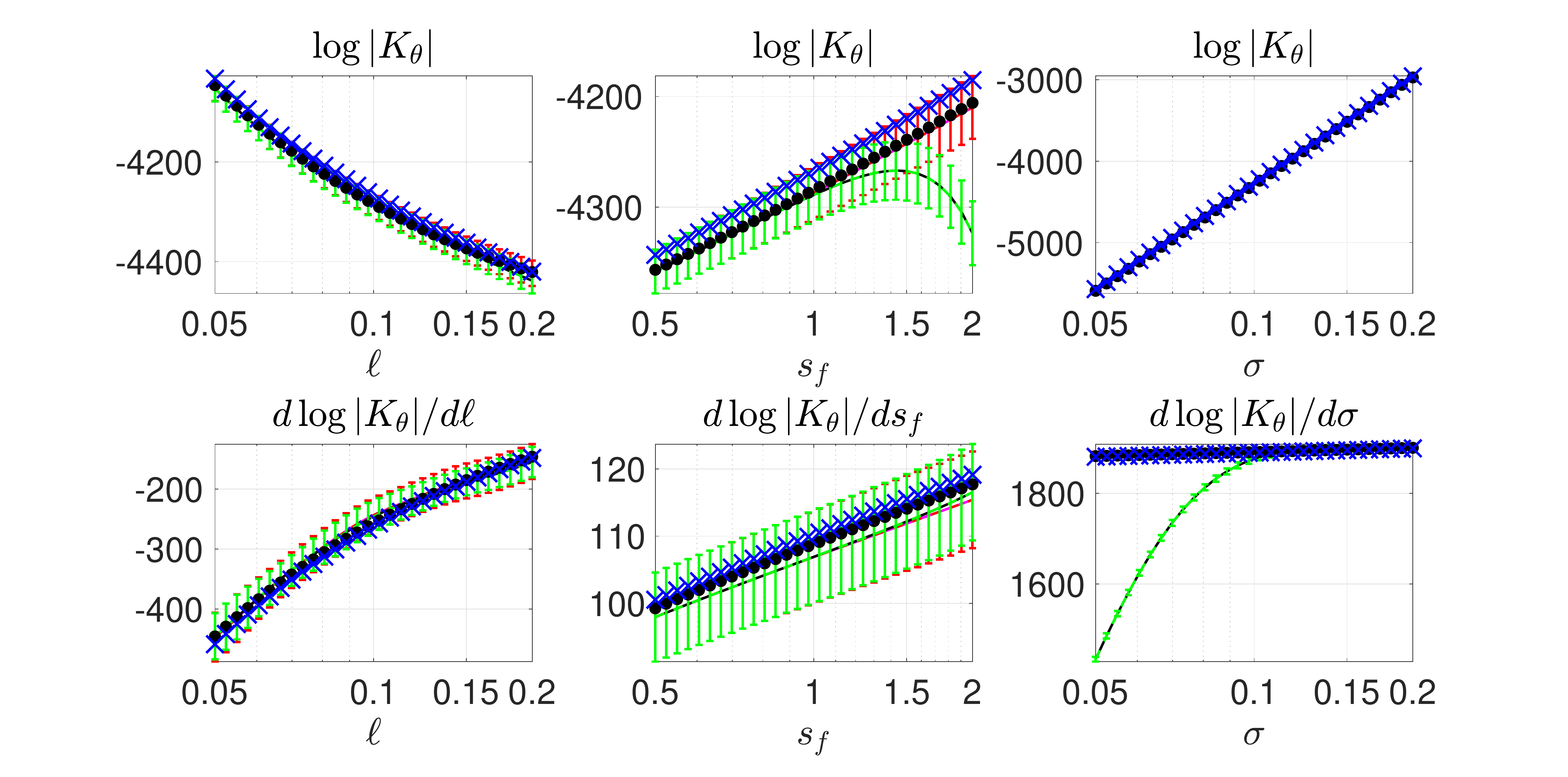}}
    \subfigure[\scriptsize log determinant for the Mat\'ern kernel]{\label{fig:1dpert_kissgp_d}
      \includegraphics[width=0.46\textwidth,trim=2.5cm 0cm 3.5cm 0cm,clip]{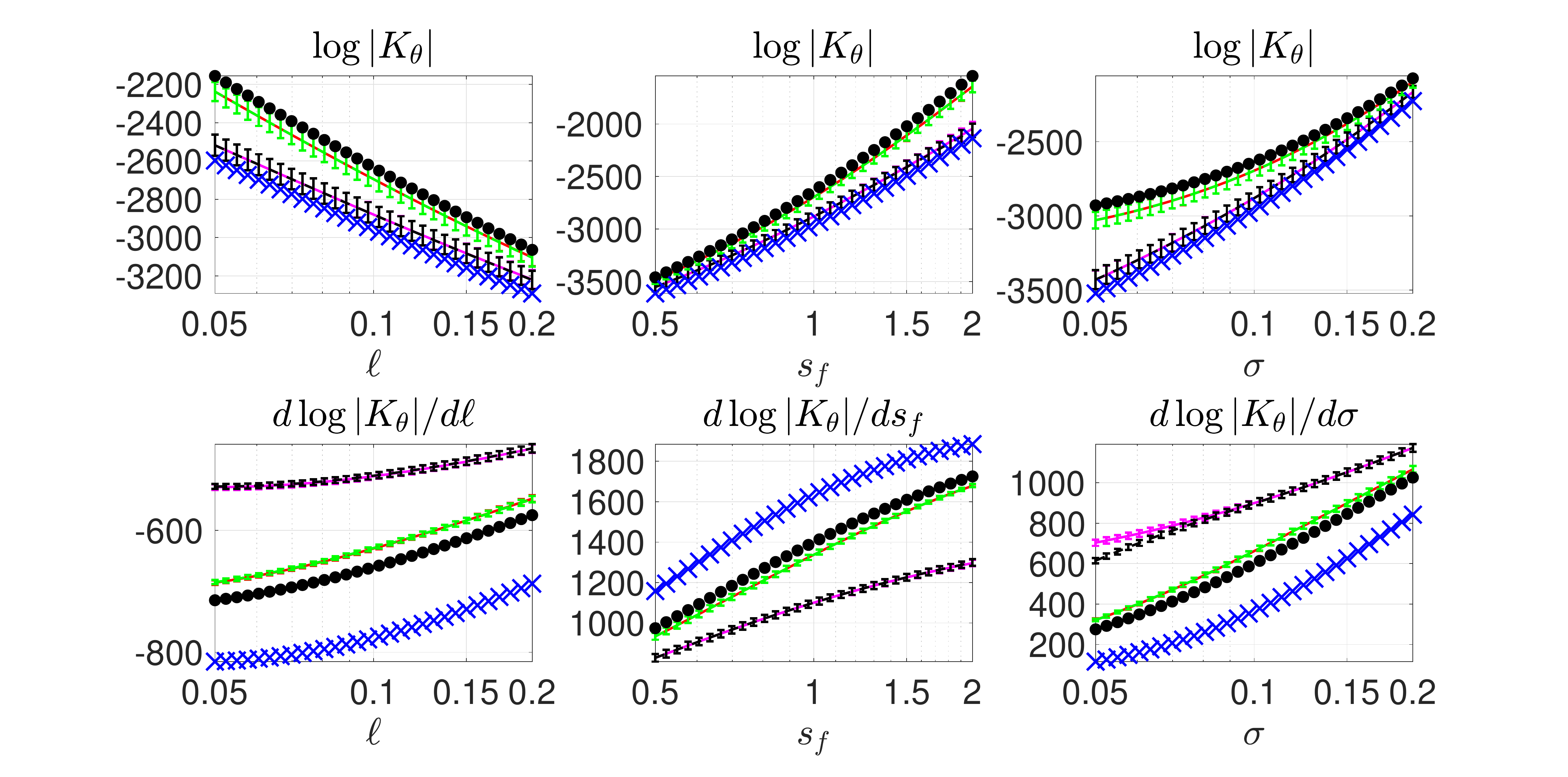}}
    \caption{1-dimensional perturbations with the SKI (cubic) approximations of the
             RBF and Mat\'ern 1/2 kernel where the data is 1000 points drawn from $\mathcal{N}(0,2)$.
             The exact values are ($\bullet$), Lanczos with diagonal replacement is ({\color{red}-----}),
             Chebyshev with diagonal replacement is ({\color{green}-----}),
             Lanczos without diagonal replacement is ({\color{magenta}-----}),
             Chebyshev without diagonal replacement is ({\color{black}-----}),
             and scaled eigenvalues is ({\color{blue} $\times$}).
             Diagonal replacement makes no perceptual difference for the RBF kernel so the lines are
             overlapping in this case. The error bars of Lanczos and Chebyshev are
             $1$ standard deviation and were computed from $10$ runs with different
             probe vectors}
     \label{fig:1dpert_kissgp}
\end{figure}

In this experiment we compare the accuracy of Lanczos and Chebyshev for
1-dimensional perturbations of a set of true hyper-parameters,
and demonstrate how critical
it is to use diagonal replacement for some approximate kernels.
We choose the true
hyper-parameters to be $(\ell,s_f,\sigma) = (0.1,1,0.1)$ and consider
two different types of datasets.  The first dataset consists of $1000$ equally
spaced points in the interval $[0,4]$ in which case the kernel matrix of a
stationary kernel is Toeplitz and we can make use of fast matrix-vector
multiplication. The second dataset consists of $1000$ data points drawn
independently from a
$U(0,4)$ distribution. We use SKI with cubic
interpolation to construct an approximate kernel based on $1000$
equally spaced points. The function values are drawn from a GP with the true
hyper-parameters, for both the true and approximate kernel.
We use $250$ iterations for Lanczos and $250$ Chebyshev moments
in order to assure convergence of both methods.
The
results for the first dataset with the RBF and Mat\'ern kernels can be seen in
Figure \ref{fig:1dpert_a}-\ref{fig:1dpert_d}. The results for the second dataset
with the SKI kernel can be seen in Figure
\ref{fig:1dpert_kissgp_a}-\ref{fig:1dpert_kissgp_d}.

Lanczos yields an excellent approximation to the log determinant and its
derivatives for both the exact and the approximate kernels, while Chebyshev
struggles with large values of $s_f$ and small values of $\sigma$ on the
exact and approximate RBF kernel. This is expected since Chebyshev has issues
with
the singularity at zero while Lanczos has large quadrature weights
close to zero to compensate for this singularity. The
scaled eigenvalue method has issues with the approximate Mat\'ern 1/2 kernel.

\subsection{Why Lanczos is better than Chebyshev}
\label{sup:lanczoscheb}

\begin{figure}[!ht]
    \centering
    \subfigure[\scriptsize True spectrum]{\label{fig:true_spectral}
        \includegraphics[width=0.46\textwidth,trim=0cm 0cm 2.5cm 0.5cm,clip]{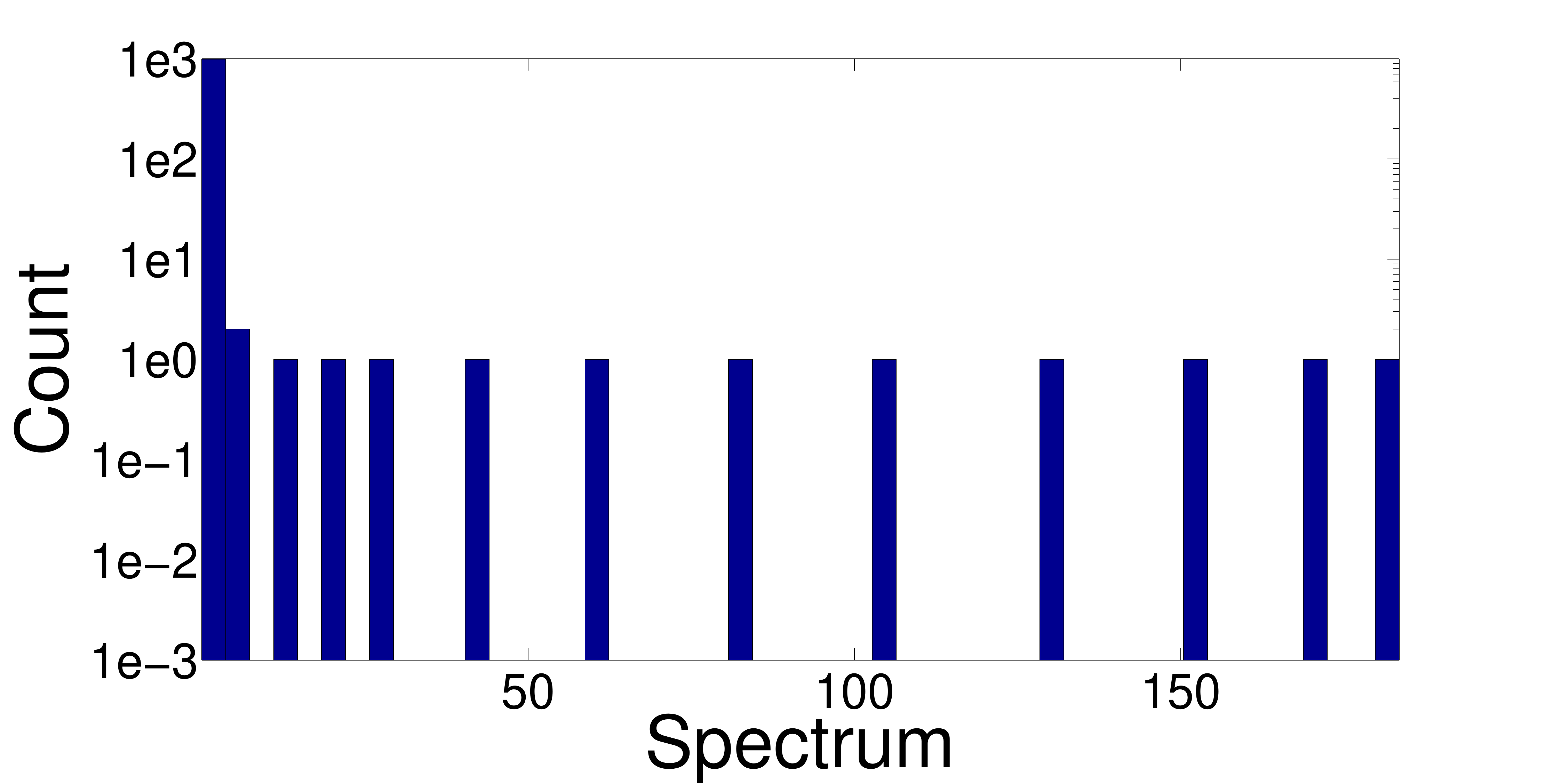}}
    \subfigure[\scriptsize Lanczos weights]{\label{fig:lanc_spectral}
        \includegraphics[width=0.46\textwidth,trim=0cm 0cm 2.5cm 0.5cm,clip]{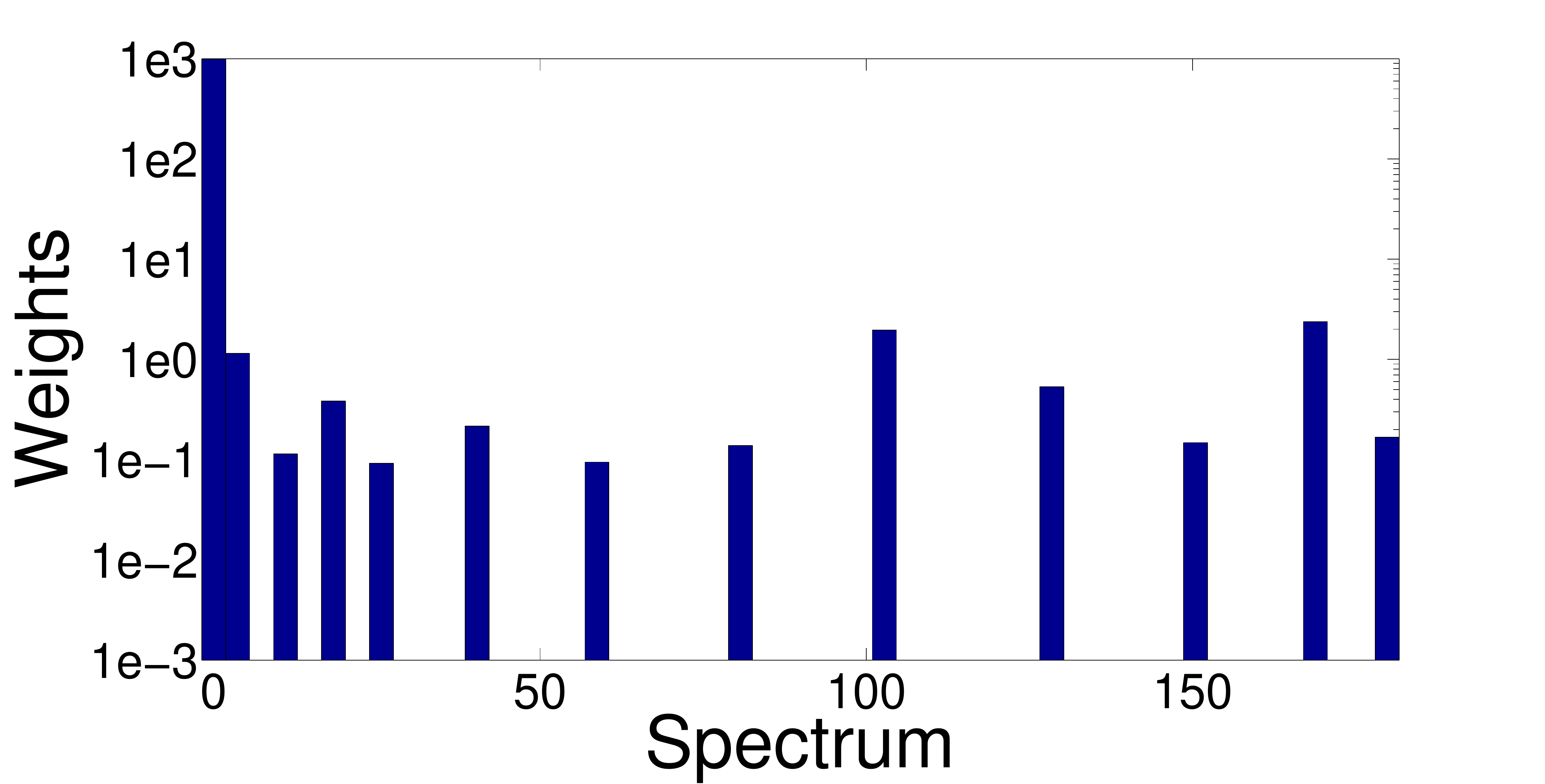}}
    \subfigure[\scriptsize Chebyshev weights]{\label{fig:cheb_spectral}
        \includegraphics[width=0.46\textwidth,trim=0cm 0cm 2.5cm 0.5cm,clip]{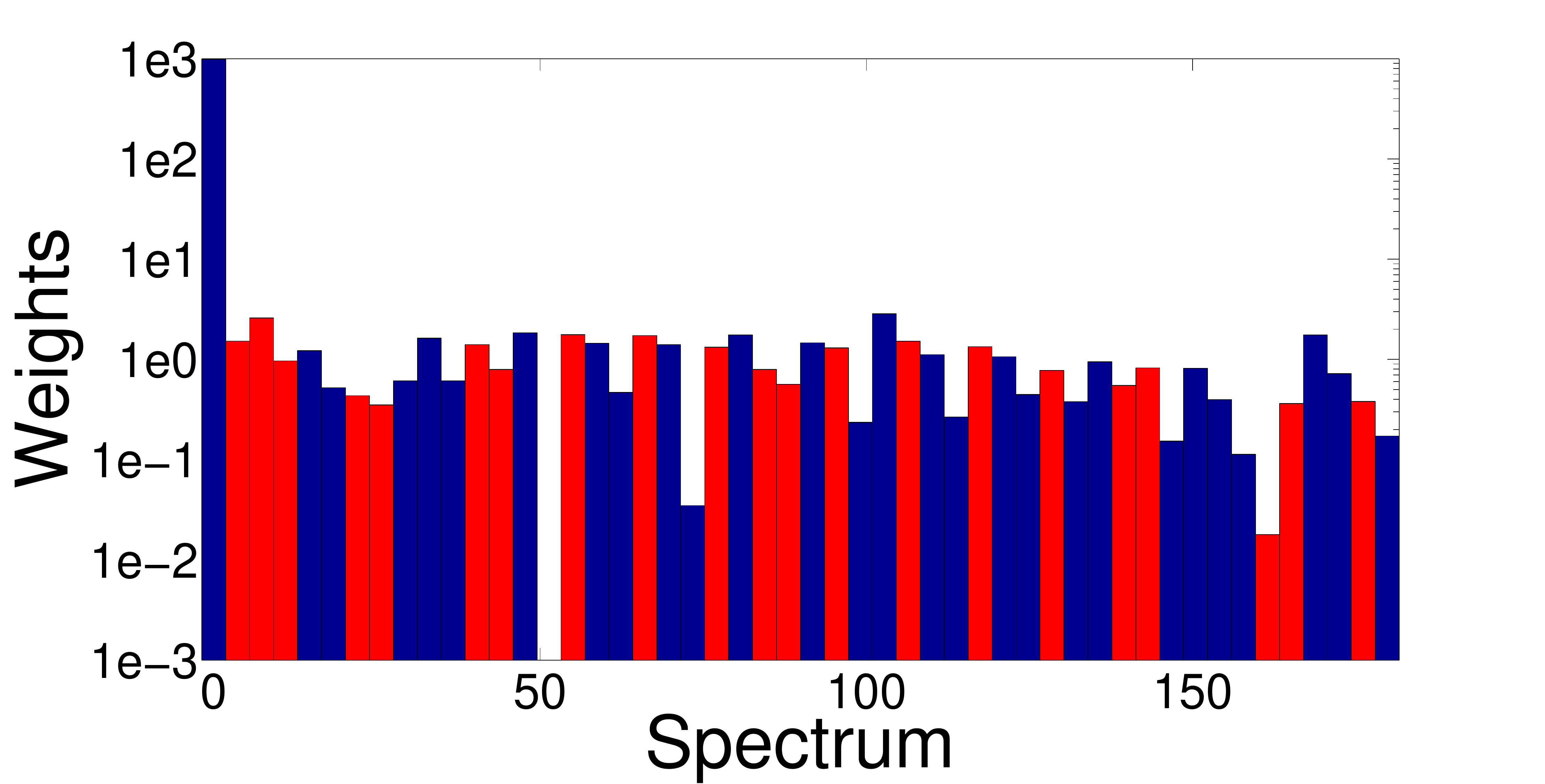}}
    \subfigure[\scriptsize Chebyshev absolute error]{\label{fig:cheb_err}
        \includegraphics[width=0.46\textwidth,trim=0cm 0cm 2.5cm 0.5cm,clip]{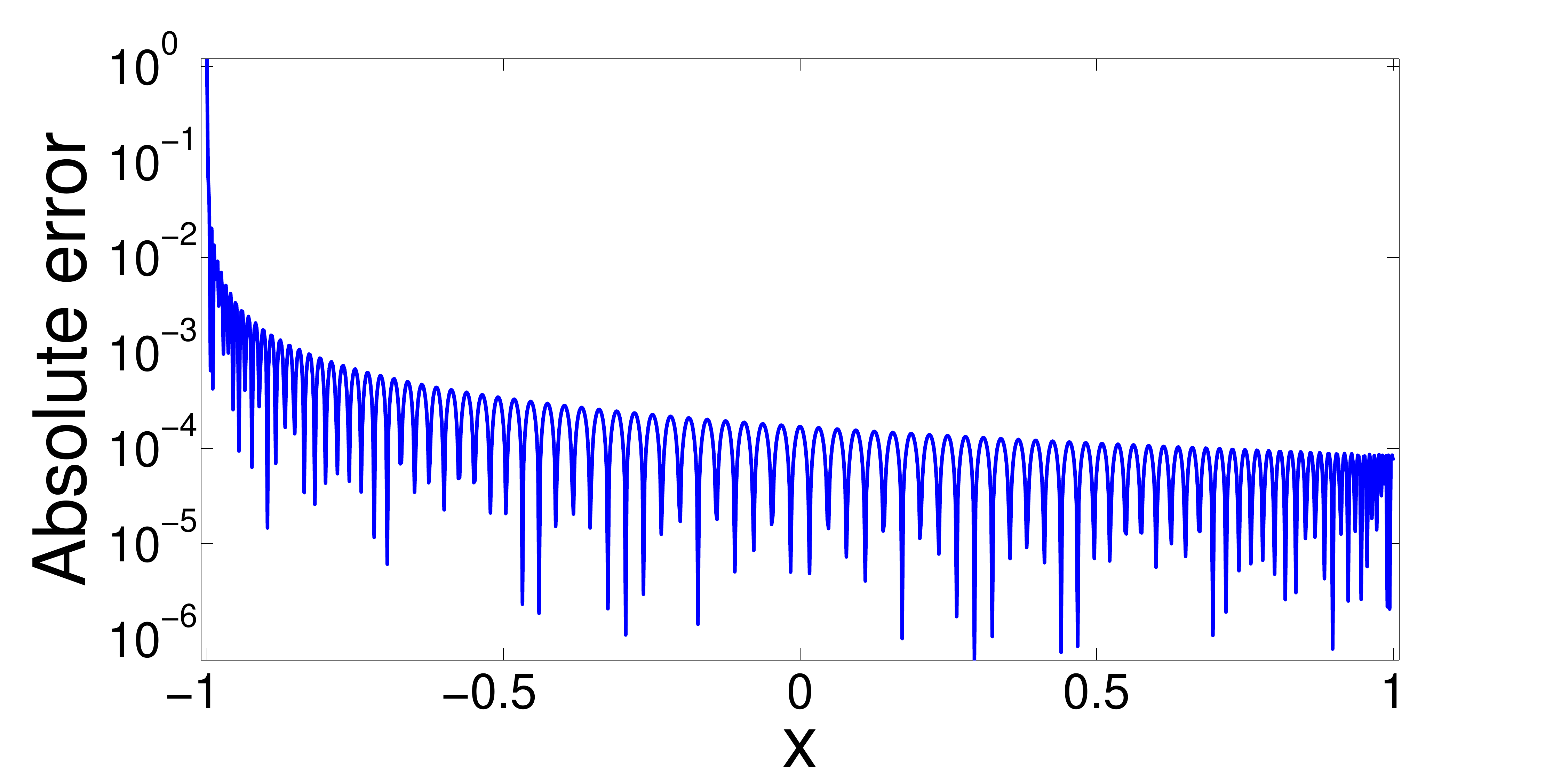}}
    \caption{A comparison between the true spectrum, the Lanczos weights ($m=50)$, and the
          Chebyshev weights ($m=100$) for the RBF kernel with $\ell=0.3$, $s_f=1$,
          and $\sigma=0.1$. All weights and counts are on a log-scale so that
          they are easier to compare. Blue bars correspond to positive weights
          while red bars correspond to negative weights.}
    \label{fig:spectrum}
\end{figure}

In this experiment, we study the performance advantage of Lanczos over Chebyshev.
Figure
\ref{fig:spectrum} shows that the Ritz values of Lanczos quickly converge
to the spectrum of the RBF kernel thanks to the absence of interior eigenvalues.
The Chebyshev approximation shows the expected equioscillation
behavior. More importantly, the Chebyshev approximation for logarithms has its
greatest error near zero where the majority of the eigenvalues are, and
those also have the heaviest weight in the log determinant.

Another advantage of Lanczos is that it requires
minimal knowledge of the spectrum, while
Chebyshev needs the extremal eigenvalues for
rescaling. In addition, with Lanczos we can get the derivatives with only one
MVM per hyper-parameter, while Chebyshev requires an MVM
at each iteration, leading to extra computation and memory usage.

\subsection{The importance of diagonal correction}
\label{sup:diagcorrectionimportance}

\begin{figure}[!ht]
    \centering
    \subfigure[\scriptsize Lanczos with diagonal correction]{\label{fig:pred_lanc_diag}
    \includegraphics[width=0.46\textwidth,trim=2.5cm 0cm 2.5cm 0.5cm,clip]{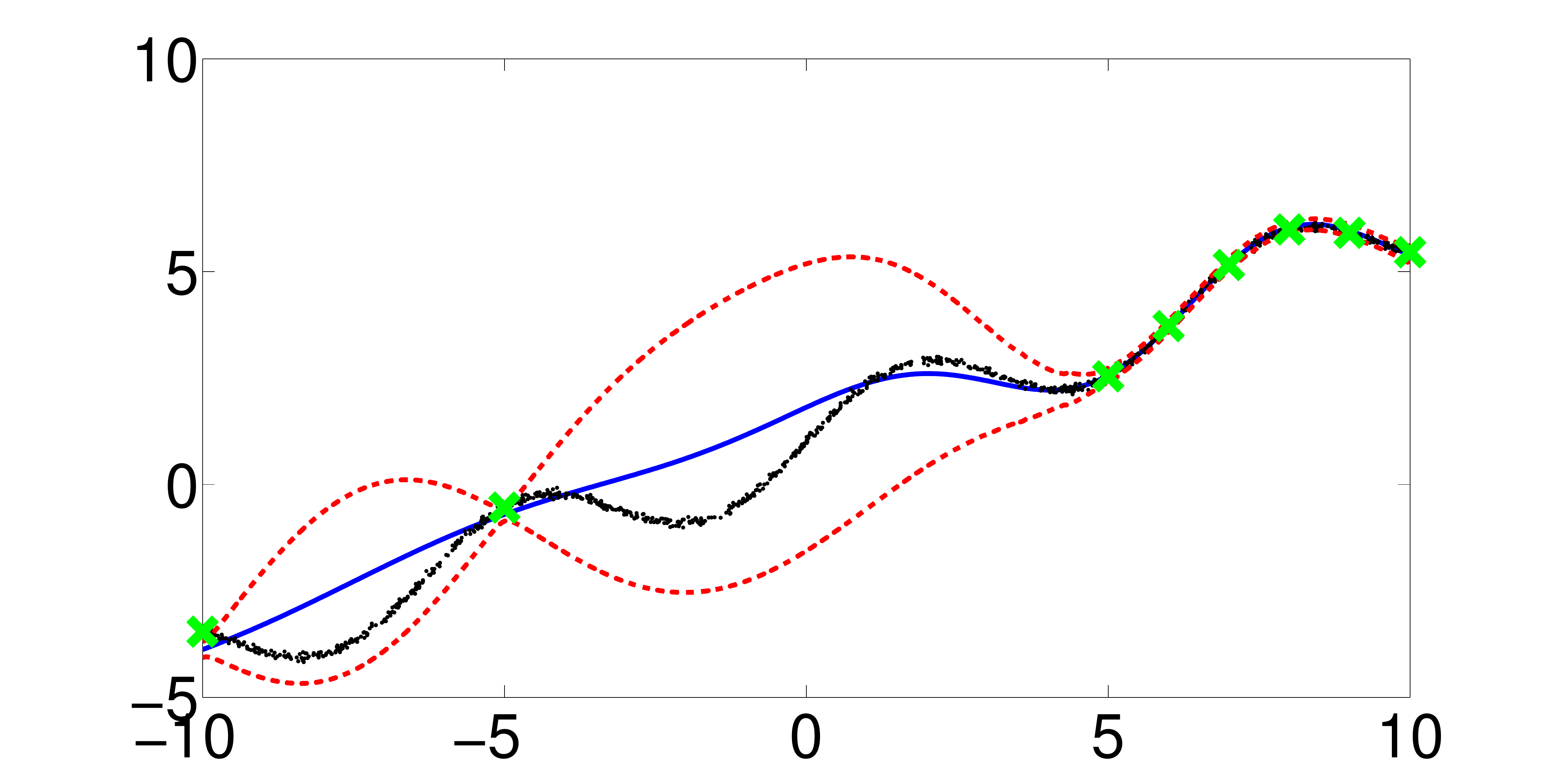}}
    \subfigure[\scriptsize Lanczos without diagonal correction]{\label{fig:pred_lanc_nodiag}
    \includegraphics[width=0.46\textwidth,trim=2.5cm 0cm 2.5cm 0.5cm,clip]{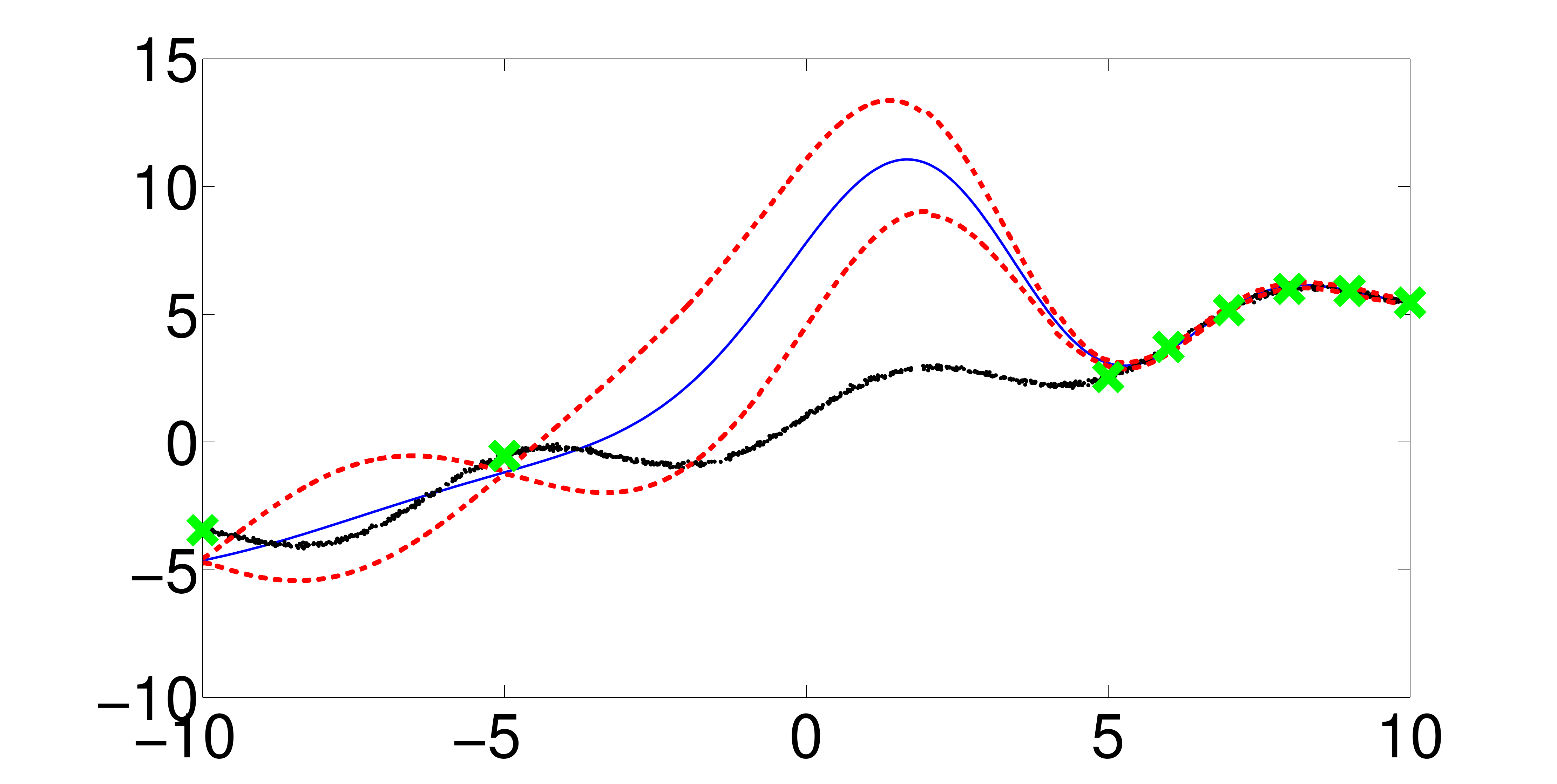}}
    \subfigure[\scriptsize Chebyshev with diagonal correction]{\label{fig:pred_cheb_diag}
    \includegraphics[width=0.46\textwidth,trim=2.5cm 0cm 2.5cm 0.5cm,clip]{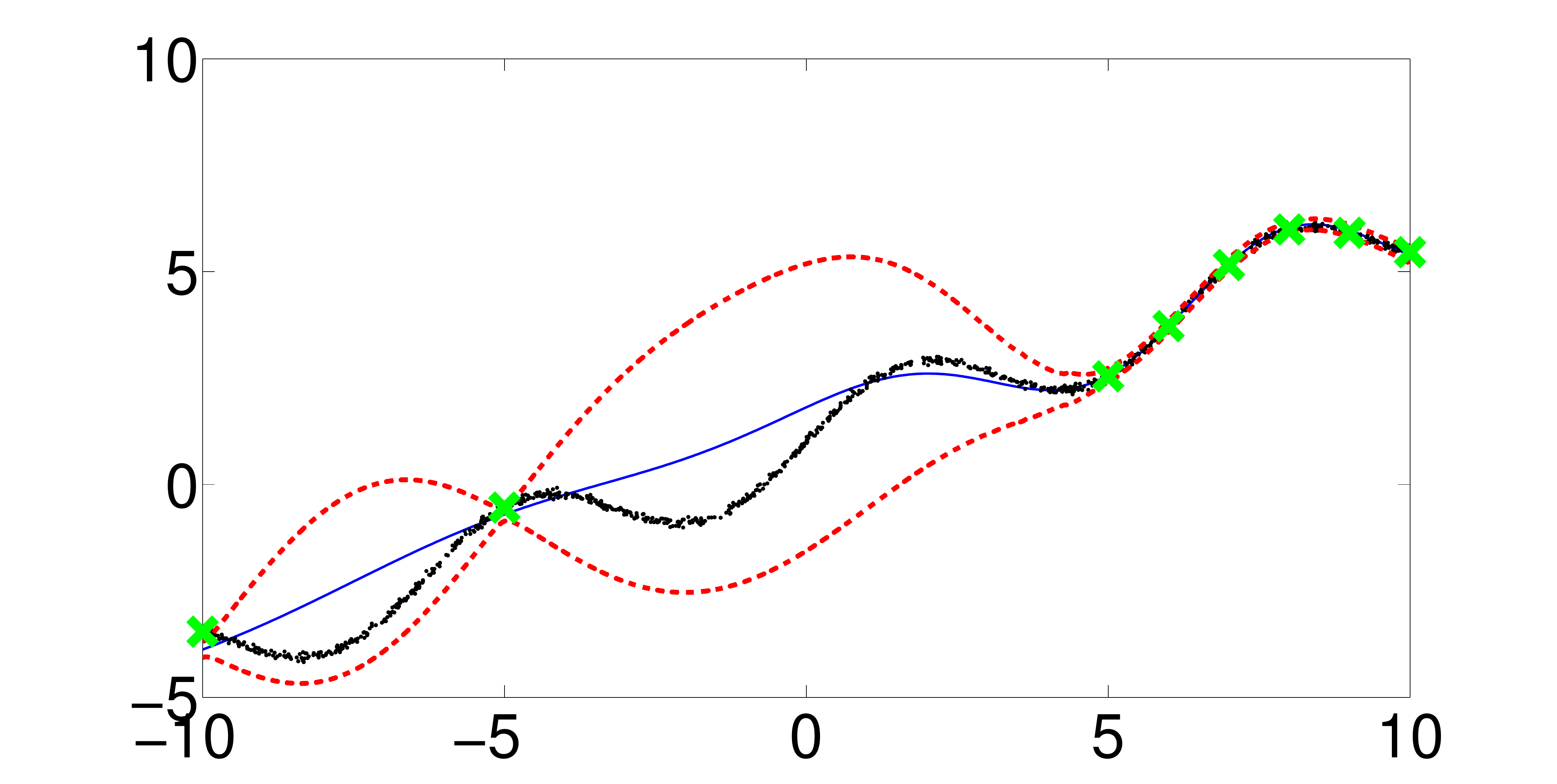}}
    \subfigure[\scriptsize Chebyshev without diagonal correction]{\label{fig:pred_lanc_nodiag}
    \includegraphics[width=0.46\textwidth,trim=2.5cm 0cm 2.5cm 0.5cm,clip]{pred_lanc_nodiag}}
    \subfigure[\scriptsize FITC]{\label{fig:pred_fitc}
    \includegraphics[width=0.46\textwidth,trim=2.5cm 0cm 2.5cm 0.5cm,clip]{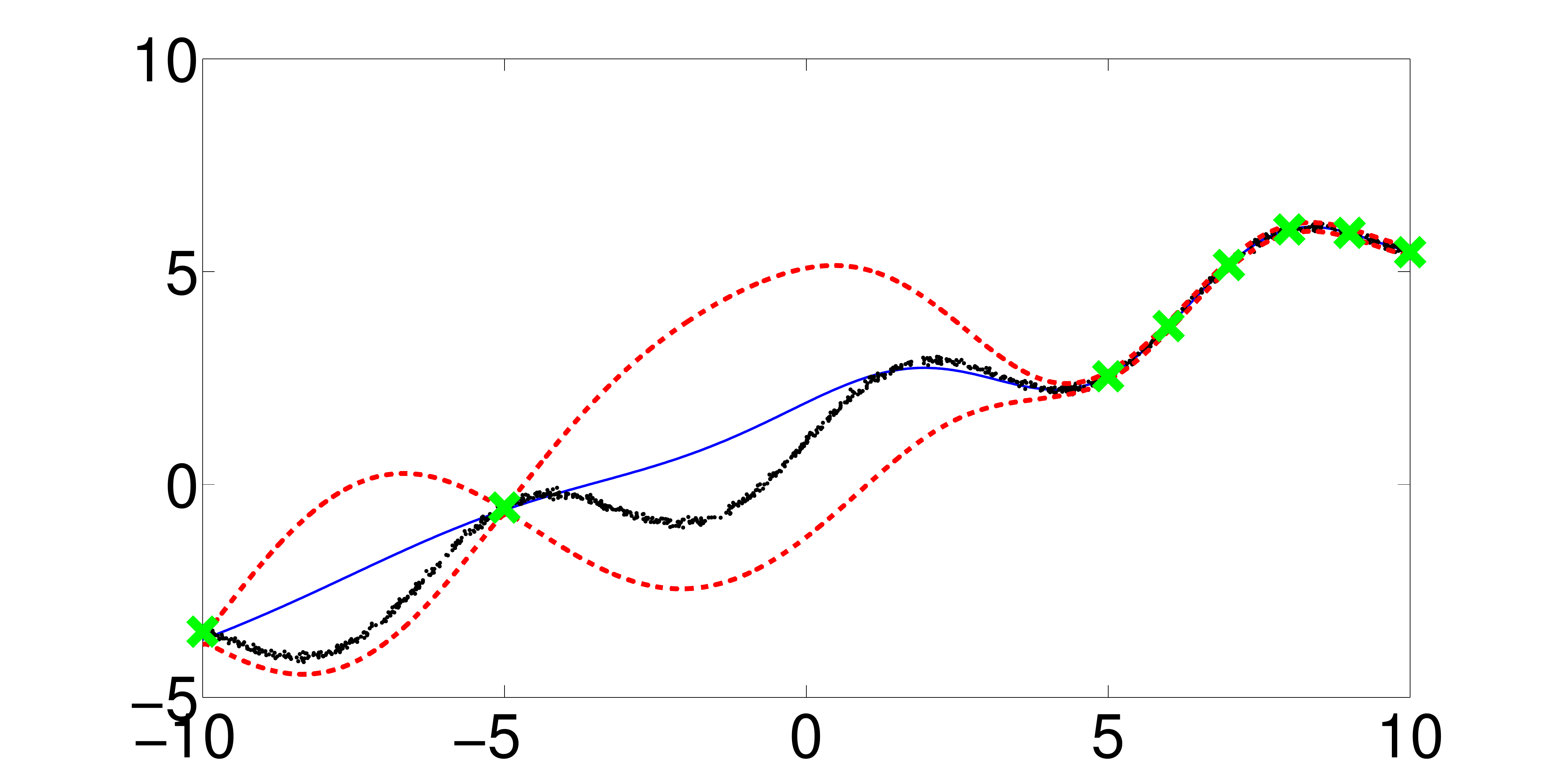}}
    \subfigure[\scriptsize Scaled eigenvalue method]{\label{fig:pred_sceig}
    \includegraphics[width=0.46\textwidth,trim=2.5cm 0cm 2.5cm 0.5cm,clip]{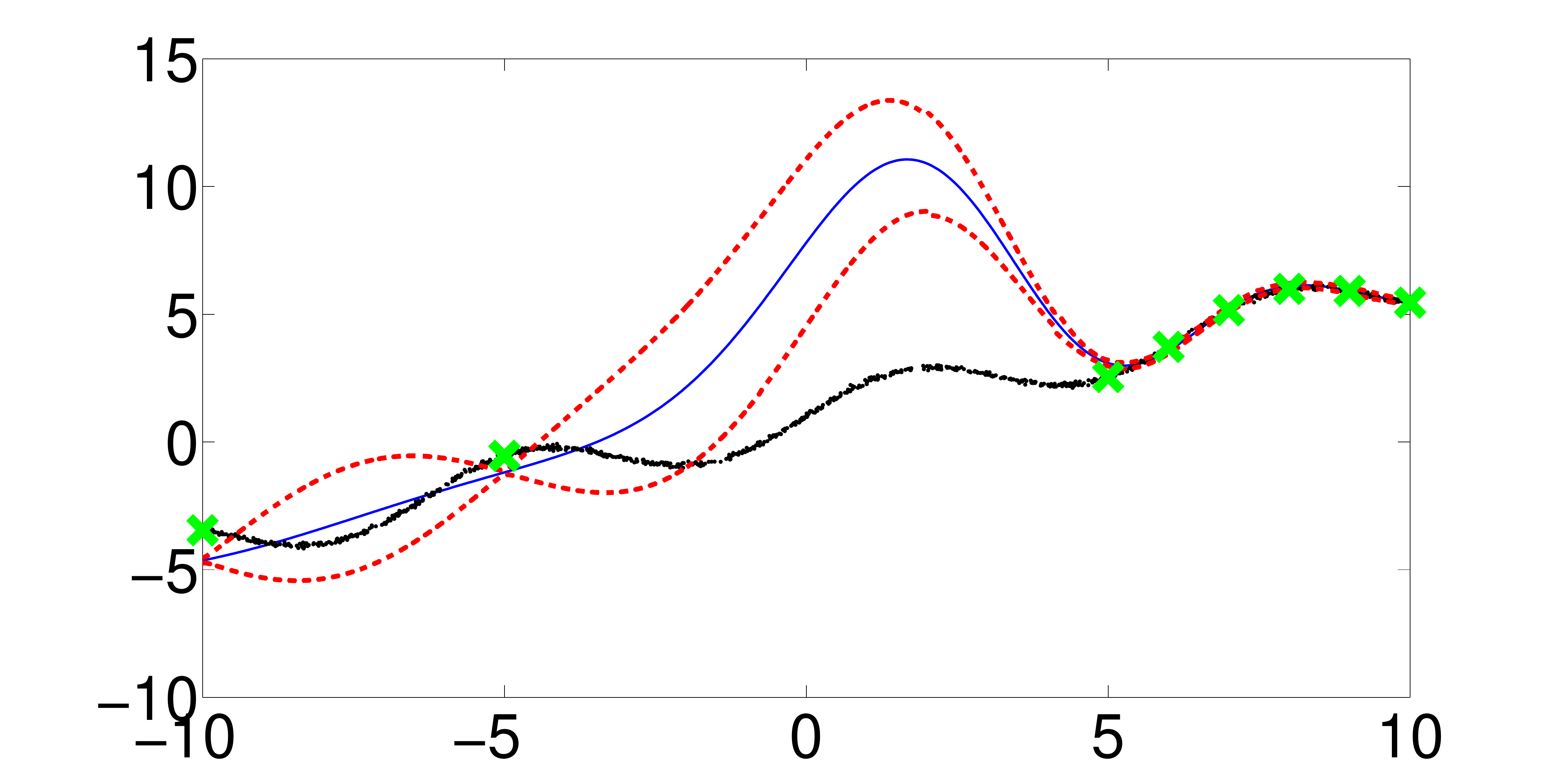}}
    \caption{Example that shows how important diagonal correction can be for
           some kernels. The Mat\'ern $3/2$ kernel was used to fit the data given
           by the black dots. This data was generated from the function
           $f(x) = 1 + x/2 + \sin(x)$ to which we added normally distributed
           noise with standard deviation $0.05$. We used the exact method to
           find the optimal hyper-parameters and used these hyper-parameters
           to study the different behavior of the predictive uncertainties
           when the inducing points are given by the green crosses. The solid
           blue line is the predictive mean and the dotted red lines shows a
           confidence interval of two standard deviations.
    }
    \label{fig:diag_correction}
\end{figure}

This experiment shows that diagonal correction of the approximate kernel
can be very important. Diagonal correction cannot be used
efficiently for some methods, such as the scaled eigenvalue method, and this
may hurt its predictive performance. Our experiment is similar to
\citep{quinonero2005unifying}.  We generate $1000$ uniformly distributed
points in the interval $[-10,10]$, and we choose a small number of inducing
points in such a way that there is a large chunk of the interval where
there is no inducing point. We are interested in the behavior of the predictive
uncertainties on this subinterval.
The function values are given by $f(x) = 1 + x/2 + \sin(x)$
and normally distributed noise with standard deviation $0.05$ is added to the
function values. We find the optimal hyper-parameters of the Mat\'ern
$3/2$ using the exact method and use these hyper-parameters to make predictions
with Lanczos, Chebyshev, FITC, and the scaled eigenvalue method. We consider Lanczos
both with and without diagonal correction in order to see how this affects
the predictions. The results can be seen in Figure \ref{fig:diag_correction}.

It is clear that Lanczos and Chebyshev are too confident in the predictive
mean when diagonal correction is not used, while the predictive uncertainties
agree well with FITC when diagonal correction is used. The scaled eigenvalue method
cannot be used efficiently with diagonal correction and we see that this leads
to predictions similar to Lanczos and Chebyshev without diagonal correction.
The flexibility of being able to use diagonal correction with Lanczos and
Chebyshev makes these approaches very appealing.

\subsection{Surrogate log determinant approximation}
\label{sup:surrlogdetapprox}

\begin{figure}[!ht]
    \centering
    \subfigure[\scriptsize RBF exact]{\label{fig:level_rbf_exact}
    \includegraphics[width=0.46\textwidth,trim=2.5cm 0cm 0.5cm 1.5cm,clip]
    {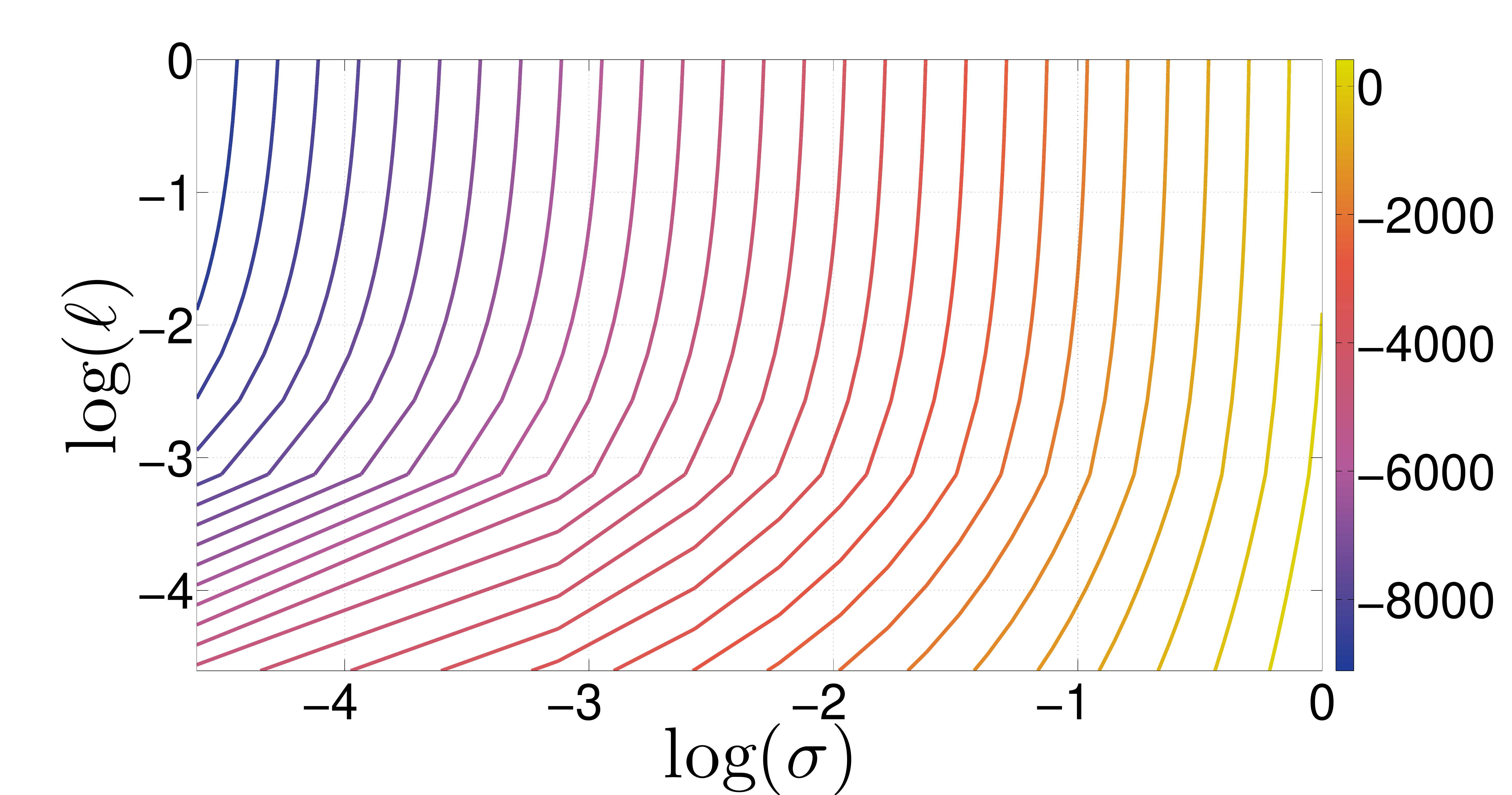}}
    \subfigure[\scriptsize Mat\'ern $3/2$ exact]{\label{fig:level_Matern3_2_exact}
    \includegraphics[width=0.46\textwidth,trim=2.5cm 0cm 0.5cm 1.5cm,clip]
    {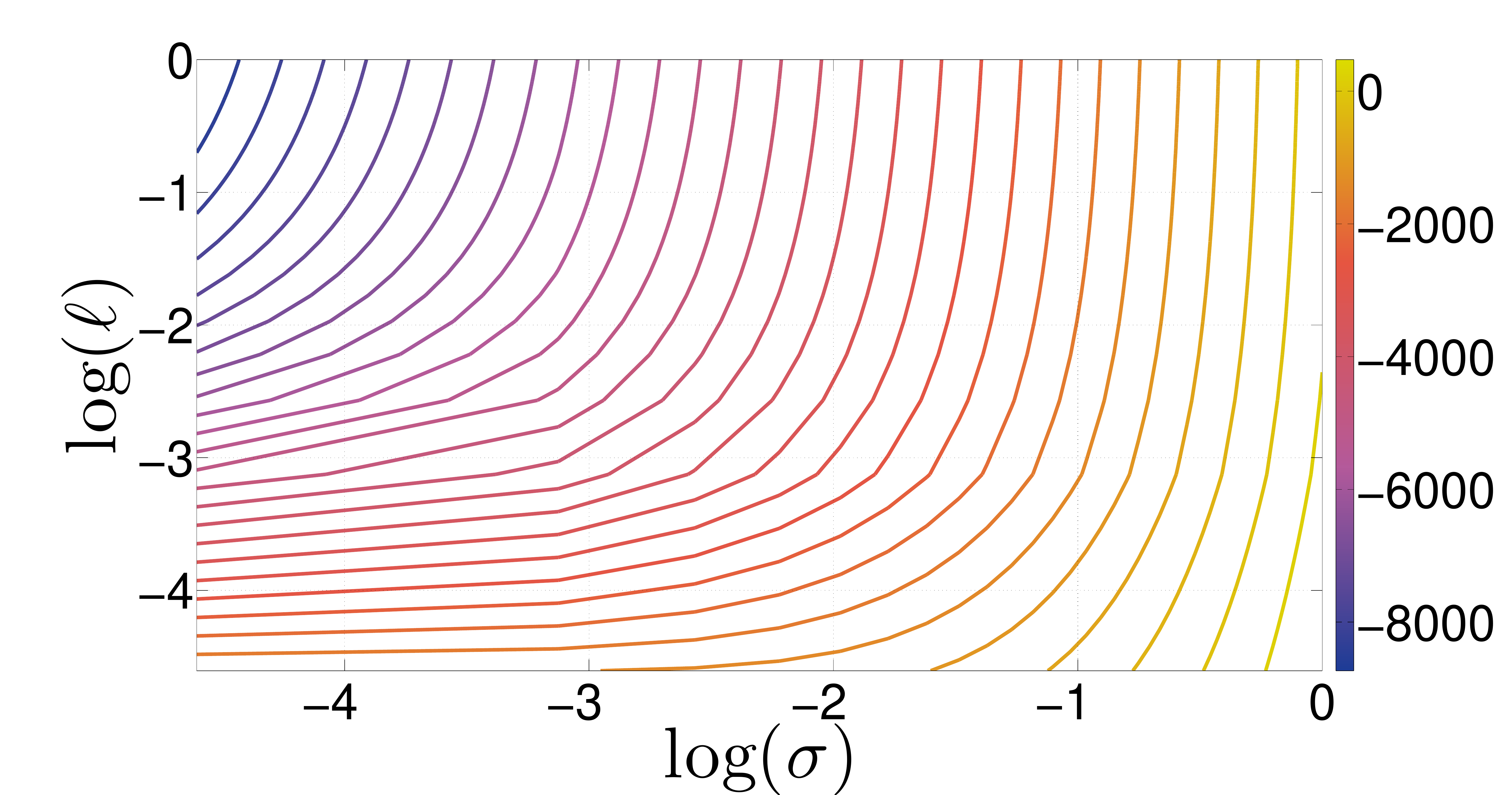}}
    \subfigure[\scriptsize RBF surrogate]{\label{fig:level_rbf_surrogate}
    \includegraphics[width=0.46\textwidth,trim=2.5cm 0cm 0.5cm 1.5cm,clip]
    {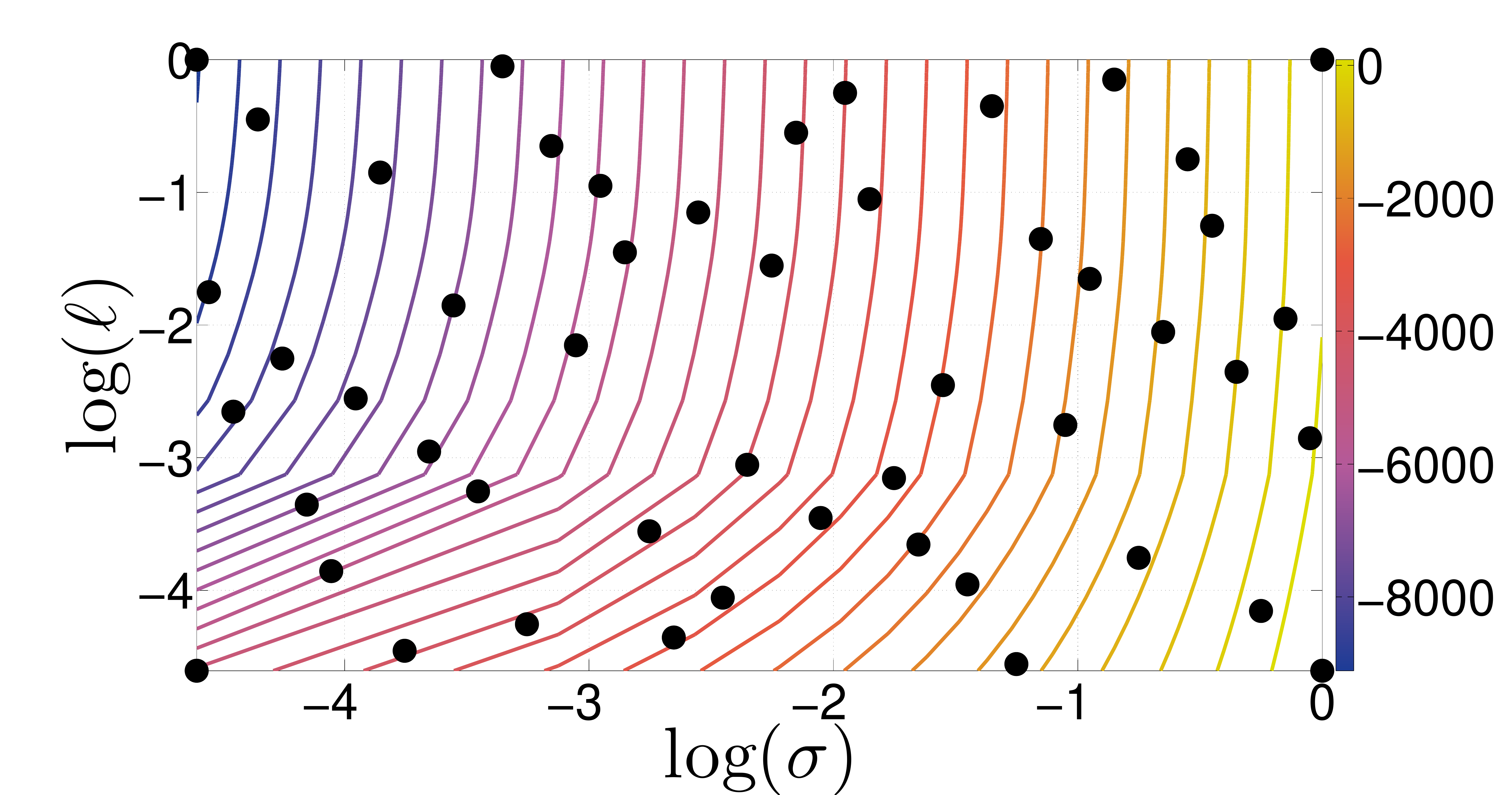}}
    \subfigure[\scriptsize Mat\'ern $3/2$ surrogate]{\label{fig:level_Matern3_2_surrogate}
    \includegraphics[width=0.46\textwidth,trim=2.5cm 0cm 0.5cm 1.5cm,clip]
    {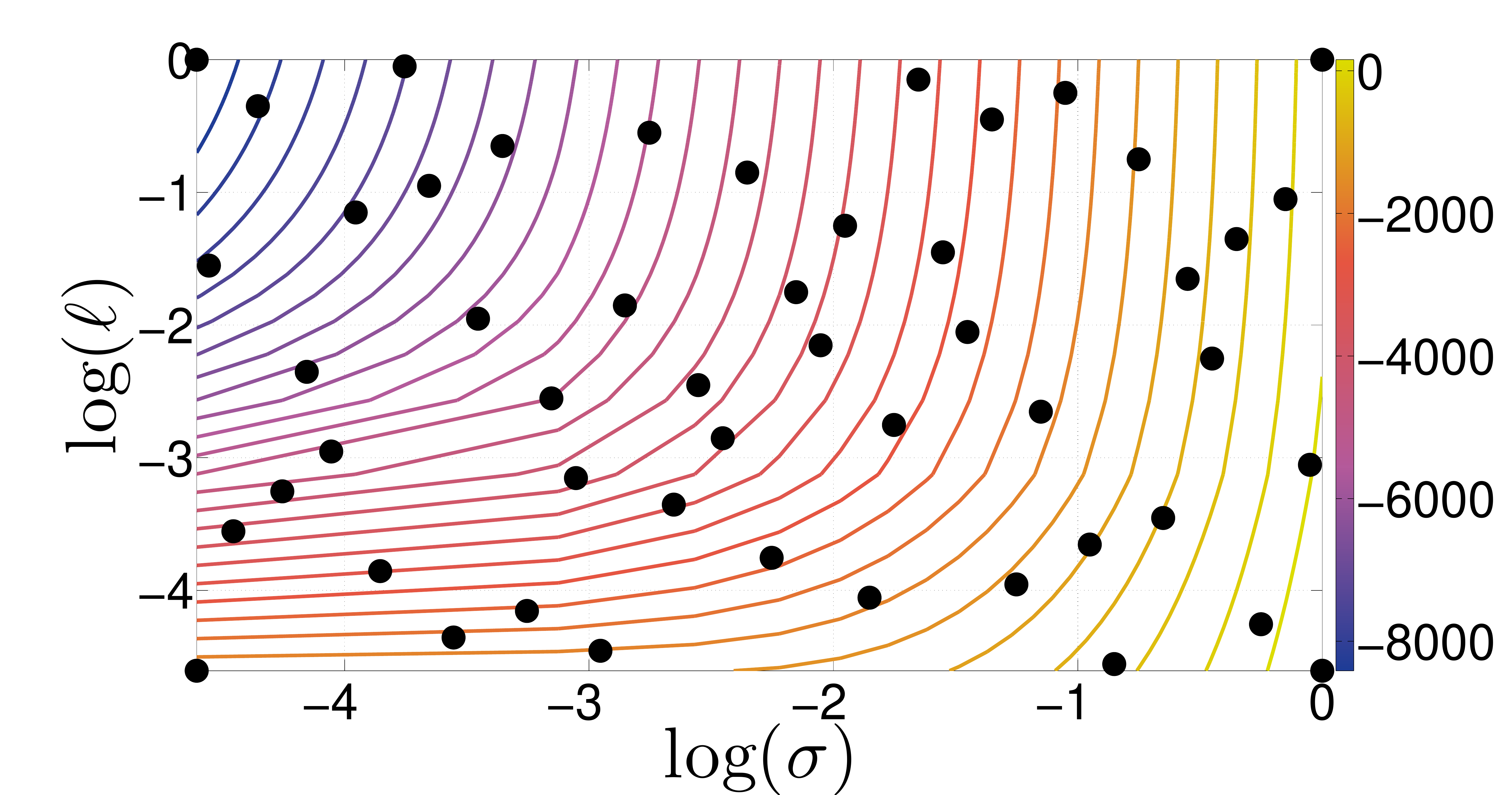}}
    \caption{Level curves of the exact and surrogate approximation of the
            log determinant as a function of $\ell$ and $\sigma$ for the RBF and
            Mat\'ern $3/2$ kernels. We used $s_f=1$ and the dataset consisted of 1000
            equally spaced points in the interval $[0,4]$. The surrogate model was
            constructed from the points shown with ($\bullet$) and the log determinant values were
            computed using stochastic Lanczos.}
\label{fig:level_curves}
\end{figure}

The point of this experiment is to illustrate how accurate the level-curves of
the surrogate model are compared to the level-curves of the true log determinant. We
consider the RBF and the Mat\'ern $3/2$ kernels and the same datasets that we
considered in \ref{sup:1dcross}. We fix $s_f=1$ and study how the level curves compare when we vary
$\ell$ and $\sigma$. Building the surrogate with all three hyper-parameters
produces similar results, but requires more design points.
We use $50$ design points to construct a cubic RBF with a linear tail.
The values of the log determinant and its derivatives
are computed with Lanczos.
It is clear from Figure \ref{fig:level_curves} that
the surrogate model does a good job approximating the log determinant for both kernels.

\subsection{Kernel hyper-parameter recovery}
\label{sup:hyperrecov}

This experiments tests how well we can recover hyper-parameters
from data generated from a GP. We compare Chebyshev, Lanczos,
the surrogate, the scaled eigenvalue method, and FITC. We consider a dataset of $5000$
points generated from a $\mathcal{N}(0,2)$ distribution. We use SKI with
cubic interpolation and a total of $2000$
inducing points for Lanczos, Chebyshev, and then scaled eigenvalue method.
FITC was used with $750$ equally spaced points because it has a longer runtime
as a function of the number of inducing points. We consider the RBF kernel and
the Mat\'ern $3/2$ kernel and sample from a GP with ground truth parameters
$(\ell,s_f,\sigma)=(0.01, 0.5, 0.05)$. The GPs for which we try to recover the
hyper-parameters were generated from the original kernel. It is important to
emphasize that there are two sources of errors present: the error from the kernel
approximation errors and the stochastic error from Lanczos and Chebyshev. We saw
in Figure \ref{fig:1dpert} and \ref{fig:1dpert_kissgp} that the stochastic error
for Lanczos is relatively small, so this follow-up experiment helps us understand
how Lanczos is influenced by the error incurred from an approximate kernel. 
We show the
true log marginal likelihood, the recovered hyper-parameters, and the run-time
in Table \ref{tab:hyper_recovery}.

\begin{table}[!ht]
    \centering
    \begin{adjustbox}{max width=\textwidth}
        \begin{tabular}{|c|c|c|c|}
          \hline
          & & RBF & Mat\'ern $3/2$ \\
          \hline
          \multirow{2}{*}{True}
          & $-\log p(y|\theta)$ & $-6.22\ee{3}$                           & $-4.91\ee{3}$     \\
          & Hypers              & $(0.01,0.5,0.05)$                       & $(0.01,0.5,0.05)$ \\
          \hline
          \multirow{3}{*}{Exact}
          & $-\log p(y|\theta)$ & $-6.23\ee{3}$                           & $-4.91\ee{3}$                           \\
          & Hypers              & $(1.01\ee{-2},4.81\ee{-1},5.03\ee{-2})$ & $(9.63\ee{-3},4.87\ee{-1},4.96\ee{-2})$ \\
          & Time (s)            & $368.9$                                 & $466.7$                                 \\
          \hline
          \multirow{3}{*}{Lanczos}
          & $-\log p(y|\theta)$ & $-6.22\ee{3}$                           & $-4.86\ee{3}$                           \\
          & Hypers              & $(1.00\ee{-2},4.77\ee{-1},5.03\ee{-2})$ & $(1.04\ee{-2},4.87\ee{-1},4.67\ee{-2})$ \\
          & Time (s)            & $66.2$                                  & $133.4$                                 \\
          \hline
          \multirow{3}{*}{Chebyshev}
          & $-\log p(y|\theta)$ & $-6.23\ee{3}$                           & $-4.81\ee{3}$                            \\
          & Hypers              & $(9.84\ee{-3},4.85\ee{-1},5.12\ee{-2})$ & $(1.11\ee{-2},4.66\ee{-1},5.78\ee{-2})$ \\
          & Time (s)            & $110.3$                                 & $173.3$                                 \\
          \hline
          \multirow{3}{*}{Surrogate}
          & $-\log p(y|\theta)$ & $-6.22\ee{3}$                           & $-4.86\ee{3}$                           \\
          & Hypers              & $(1.01\ee{-2},4.88\ee{-1},4.85\ee{-2})$ & $(1.02\ee{-2},4.80\ee{-1},4.66\ee{-2})$ \\
          & Time (s)            & $48.2$                                  & $44.3$                                  \\
          \hline
          \multirow{3}{*}{Scaled eigenvalues}
          & $-\log p(y|\theta)$ & $-6.22\ee{3}$                           & $-4.71\ee{3}$                           \\
          & Hypers              & $(1.04\ee{-2},4.52\ee{-1},5.14\ee{-2})$ & $(1.13\ee{-2},4.53\ee{-1},6.37\ee{-2})$ \\
          & Time (s)            & $90.2$                                  & $127.3$                                 \\
          \hline
          \multirow{3}{*}{FITC}
          & $-\log p(y|\theta)$ & $-6.22\ee{3}$                           & $-4.11\ee{3}$                           \\
          & Hypers              & $(1.03\ee{-2},4.90\ee{-1},5.07\ee{-2})$ & $(1.34\ee{-2},5.22\ee{-1},8.91\ee{-2})$ \\
          & Time (s)            & $86.6$                                  & $136.9$                                 \\
          \hline
      \end{tabular}
  \end{adjustbox}
  \caption{Hyper-parameter recovery for the RBF and Mat\'ern $3/2$ kernels. The data
           was generated from $5000$ normally distributed points. Lanczos,
           surrogate, and scaled eigenvalues all used 2000 inducing points while
           FITC used 750. These numbers where chosen
           to make their run times close to equal. Diagonal correction was
           applied to the Mat\'ern $3/2$
           approximate kernel. The value of the log marginal likelihood was
           was computed from the exact kernel and shows the value of the
           hyper-parameters recovered by each method. We ran Lanczos 5 times
           and averaged the values.
            }
    \label{tab:hyper_recovery}
\end{table}

It is clear from Table \ref{tab:hyper_recovery} that most methods are able to
recover parameters close to the ground truth for the RBF kernel. The results
are more interesting for the Mat\'ern $3/2$ kernel where FITC struggles and the
parameters recovered by FITC have a value of the log marginal likelihood that
is much worse than the other methods.

\end{document}